\PassOptionsToPackage{table,xcdraw}{xcolor}
\documentclass{article}

\usepackage[preprint]{neurips_2025}

\usepackage{makecell}
\usepackage{tabularx}     % new
\usepackage{adjustbox}
\usepackage{rotating}
\usepackage[utf8]{inputenc} % allow utf-8 input
\usepackage[T1]{fontenc}    % use 8-bit T1 fonts
\usepackage{hyperref}       % hyperlinks
\usepackage{url}            % simple URL typesetting
\usepackage{booktabs}       % professional-quality tables
\usepackage{amsfonts}       % blackboard math symbols
\usepackage{nicefrac}       % compact symbols for 1/2, etc.
\usepackage{microtype}      % microtypography
\usepackage{enumitem}
\usepackage{amsmath}
\usepackage{graphicx}
\usepackage{multirow} % in preamble
\usepackage{wrapfig}
\usepackage{lipsum} % For sample text
\usepackage{caption}  % in your preamble
\usepackage{csquotes} % nice and robust quotations with \enquote{}
\usepackage{pdflscape}
\usepackage{amssymb}
\usepackage{subcaption}
\usepackage{float}  % for [H] option
\usepackage{tikz}
\usepackage[table,xcdraw]{xcolor}
\usepackage[most]{tcolorbox}

\newcommand*\circled[1]{\tikz[baseline=(char.base)]{
            \node[shape=circle,draw,inner sep=0.5pt] (char) {#1};}}

\tcbset{
  mysummarybox/.style={
    enhanced,
    colback=gray!10, % light gray background
    colframe=black, % black frame
    coltitle=white, % black title color
    boxrule=1pt, % thickness of the frame
    left=1mm, % adjust the left margin
    right=1mm, % adjust the right margin
    top=1mm, % adjust the top margin
    bottom=1mm, % adjust the bottom margin
    arc=2mm % adjust the roundness of the corners
  }
}

\title{A Guide to Robust Generalization: The Impact of Architecture, Pre-training, and Optimization Strategy}

\author{
  \textbf{Maxime Heuillet}$^{1,2}$ \quad
  \textbf{Rishika Bhagwatkar}$^{3,2}$ \quad
  \textbf{Jonas Ngnawé}$^{1,2}$ \quad
  \textbf{Yann Pequignot}$^{1}$ \\
  \textbf{Alexandre Larouche}$^{1,2}$ \quad
  \textbf{Christian Gagné}$^{4,1,2}$ \quad
  \textbf{Irina Rish}$^{4,3,2}$ \\
  \textbf{Ola Ahmad}$^{5}$ \quad
  \textbf{Audrey Durand}$^{4,1,2}$ \\
  \\
  $^{1}$Université Laval (IID) \quad
  $^{2}$Mila - Québec AI Institute \quad
  $^{3}$Université de Montréal \\
  $^{4}$Canada CIFAR AI Chair - Mila \quad
  $^{5}$CortAIx labs, Thales Digital Solutions
}

\newif\ifdraft
\drafttrue  % Set to \draftfalse before submission

\begin{document}

\maketitle

\begin{abstract}
Deep learning models operating in the image domain are vulnerable to small input perturbations.
For years, robustness to such perturbations was pursued by training models from scratch (i.e., with random initializations) using specialized loss objectives.
Recently, robust fine-tuning has emerged as a more efficient alternative: instead of training from scratch, pretrained models are adapted to maximize predictive performance and robustness. 
To conduct robust fine-tuning, practitioners design an optimization strategy that includes the model update protocol (e.g., full or partial) and the specialized loss objective.
Additional design choices include the architecture type and size, and the pretrained representation.
These design choices affect robust generalization, which is the model’s ability to maintain performance when exposed to new and unseen perturbations at test time.
Understanding how these design choices influence generalization remains an open question with significant practical implications.
In response, we present an empirical study spanning $6$ datasets, $40$ pretrained architectures, $2$ specialized losses, and $3$ adaptation protocols — yielding $1,440$ training configurations and $7,200$ robustness measurements across five perturbation types. 
To our knowledge, this is the most diverse and comprehensive benchmark of robust fine-tuning to date. 
While attention-based architectures and robust pretrained representations are increasingly popular, we find that convolutional neural networks pretrained in a supervised manner on large datasets often perform best. 
Our analysis both confirms and challenges prior design assumptions, highlighting promising research directions and offering practical guidance.
\end{abstract}

\section{Introduction}

% AT can be used on any architecture \citep{singh2024revisiting} and across modalities \citep{xhonneux2024efficient,gan2020large}. 
% In addition to the fine-tuning  to choose a specialized optimization technique to induce robustness.
% The most effective technique is \textit{adversarial training} (AT) \citep{madry2017towards}, which consists in minimizing a loss objective on training data that is synthetically perturbed.
%To motivate this hypothesis, we illustrate in Figure \ref{fig:perf_improvment} the important performance variability across design choices. 
% Despite attempts to decrease AT's computational costs \citep{shafahi2019adversarial, zhang2022revisiting}, adversarial training is more computationally intensive than standard (non-robust) training \citep{bartoldson2024adversarial}.
%\ola{... to their deployment in real-world.}
% The optimization strategy includes the design of the fine-tuning protocol, and the choice of a specialized objective to induce robustness.

Images processed by machine learning models can contain subtle perturbations that are invisible to the human eye.
These perturbations may occur accidentally (e.g. sensor noise, blur, digital format conversions \citep{imgaug}) or intentionally (e.g., adversarial attacks \citep{szegedy2013intriguing}).
Such perturbations can negatively affect the performance of machine learning systems, which is a serious obstacle to their adoption in the real world.

In practice, it is difficult to anticipate which type(s) of perturbation(s) a system may face \citep{sculley2015hidden}. 
A key challenge is therefore to maximize robustness across diverse perturbation types.
To achieve that, a typical approach is to assume a set of possible perturbations and induce robustness to this specific set during training \citep{croce2022adversarial, tramèr2019adversarialtrainingrobustnessmultiple, maini2020adversarial}.
However, this strategy is inherently limited, as models may encounter unforeseen perturbations post-deployment \citep{bashivan2021adversarial, ibrahim2022towards}. 
In this work, we focus on \textit{robust generalization}: it refers to the ability of models trained for robustness on a specific perturbation type to remain robust to other, unseen, perturbations.
% it refers to the robustness of models trained on specific perturbation types to other, unseen, perturbations types.
%we adapt models for robustness against a specific type of perturbation and investigate how well this induces robustness to other, unseen perturbations. %adversarial

% We specifically focus on robust generalization in low data regimes. %, where training datasets include around $10$k observations. 
The experimental conditions of this study focus on robust image classification in the low training data regime. The experimental conditions of this study represent a frequent problem faced by practitioners in diverse applications (e.g. aerospace \citep{deng2018hyperspectral}, medical imaging \citep{alzubaidi2021novel,oh2020deep}).
% Robustness-critical applications often face data scarcity constraints, due to data collection costs \citep{rahimi2021addressing}.
In low data regimes, robust generalization can be induced by fine-tuning for robustness models pre-trained on large datasets \citep{hua2024initialization,xu2023autolora,hendrycks2019using, liu2023twins}.
% To improve robust generalization in the low data regime, practitioners fine-tune for robustness models that have been pre-trained on large datasets \citep{hua2024initialization,xu2023autolora,hendrycks2019using, liu2023twins}.

Fine-tuning for robustness involves a wide range of design choices related to the pretrained backbone and the fine-tuning process.
% However, fine-tuning for robustness involves a wide range of design choices and there is limited guidance available for practitioners to navigate these choices effectively.
When selecting a pretrained backbone, one implicitly selects an architecture type (e.g., convolutional, attention-based, or hybrid), a model size, and a pretraining strategy (e.g., supervised vs. self-supervised, robust vs. non-robust). As for the robust fine-tuning process, one must select a fine-tuning protocol (e.g., partial vs full updates) and a loss objective.
% A first step is to specify the optimization strategy that includes the model update protocol (e.g., partial, or full updates) and the loss objective.
A standard loss objective is classic adversarial training (Classic AT) \citep{madry2017towards}, which minimizes cross-entropy on adversarially perturbed observations. 
Another option is the so-called TRADES (TRadeoff-inspired Adversarial DEfense via Surrogate-loss minimization) \citep{zhang2019theoretically} loss, which optimizes both the cross-entropy and the Kullback-Leibler (KL) divergence between predictions on perturbed and unperturbed observations.
% Another option is the so-called TRADES (TRadeoff-inspired Adversarial DEfense via Surrogate-loss minimization) \citep{zhang2019theoretically} loss, which achieves robustness by adding to the cross-entropy a KL-divergence term.
% \aud{KL to minimize distance between what and what? Any expected benefit of TRADES vs Classic AT? Is TRADES more/less computationally expensive than Classic AT?} \maxi{What about: There is no consensus as to which loss to choose to perform robust fine-tuning, as suggested by inconsistent design decisions in the literature (e.g., Classic AT in \cite{hua2024initialization, singh2024revisiting}, TRADES in \cite{xu2023autolora}).}
Unfortunately, there is currently limited guidance available for practitioners to navigate these choices effectively.
% Additionally, practitioners must choose the architecture size and type (e.g., convolutional, based on attention layers or hybrid), as well as a pre-trained backbone (e.g., obtained with supervised, self-supervised or robust pre-training, etc).

\begin{wrapfigure}[15]{r}{0.48\textwidth}
  \centering
  \vspace{-18pt} % Optional: tighten vertical spacing
  \includegraphics[width=0.46\textwidth]{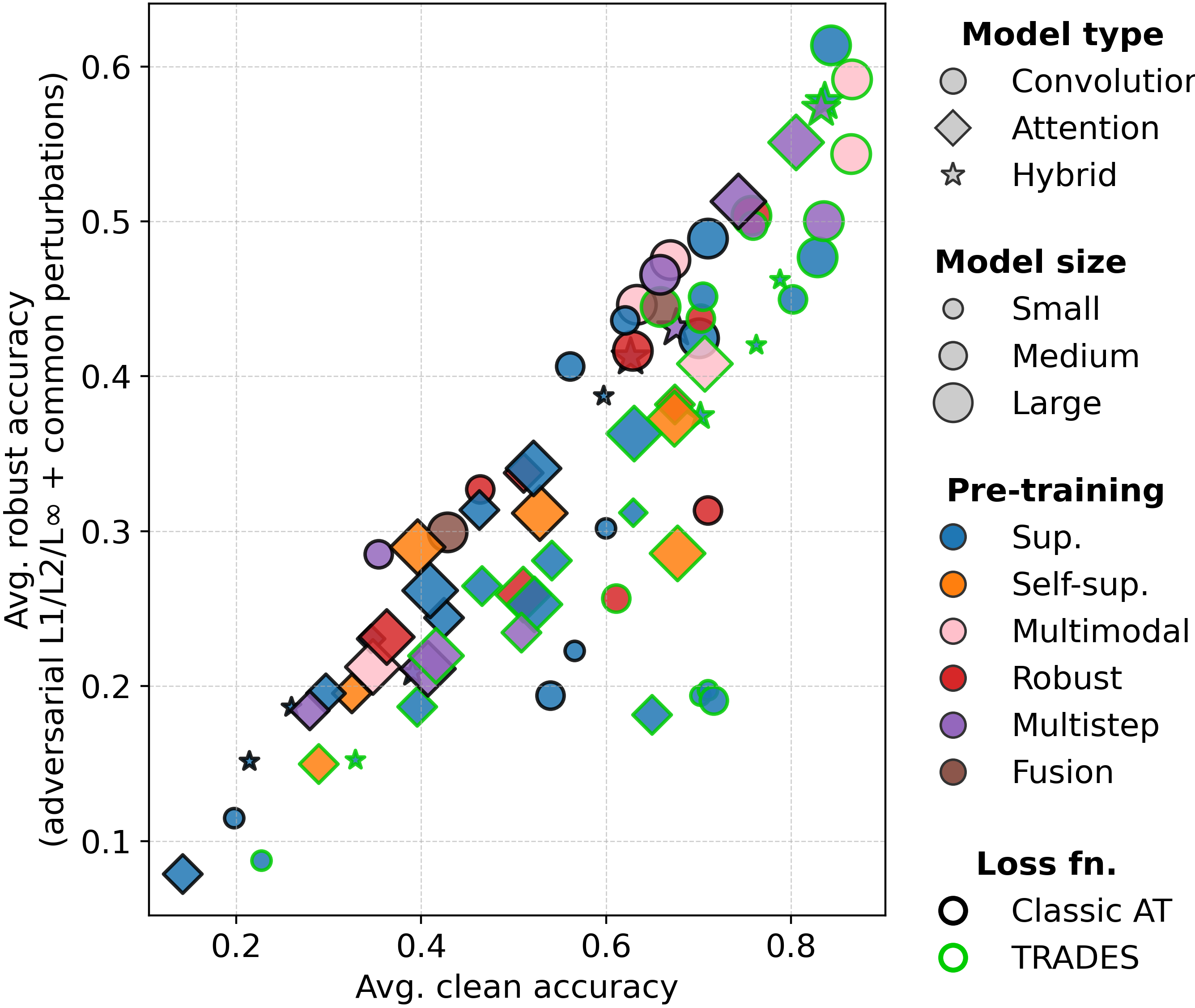}
  \caption{Performance variation across fine-tuning design choices (full fine tuning with 50 epochs). Accuracy averaged over $6$ datasets.}
  \label{fig:perf_improvment}
  \vspace{-10pt} % Optional: adjust bottom spacing
\end{wrapfigure}

\paragraph{Research question} \textit{What are the impacts of robust fine-tuning design choices on robust generalization?}
% \textit{To what extent fine-tuning for robustness with respect to a single type of perturbation, lead to machine learning models that generalize to unseen perturbations at test time?} 
Our main hypothesis is that the pretrained backbone interacts with the fine-tuning and optimization strategies to substantially influence robust generalization.
% Our main hypothesis is that \textit{design choices} such as the fine-tuning strategy, the architecture type and size,  the pre-trained backbone, and the loss objective,  and the optimization strategy, interact together to substantially influence robust generalization. 
Figure \ref{fig:perf_improvment} motivates this hypothesis by showing important performance variability across design choices and complex interaction patterns among design components.

\paragraph{Key findings}
We conduct a study on $6$ datasets and a total of $240$ design choices combinations (40 pretrained backbones $\times$ 2 robust losses $\times$ 3 fine-tuning protocols).
% \aud{According to the previous sentence it seems that the definition of ``backbone'' includes an architecture type+size and a pretraining strategy; is it right?} \maxi{yes} 
We obtain $7,200$ measurements of robustness  on 5 perturbation types. We uncover actionable lessons for practitioners and for future research on robust fine-tuning:

%\aud{What are tasks here? Datasets? Also, by ``perform best'', you mean better robust generalization?}
% compared to other pre-training strategies
%\aud{Fro point 7, I asume that this applies to small-med architectures?}
% \maxi{To validate:}
\circled{1} TRADES loss performs better than Classic AT overall and significantly better in large models. \circled{2} Despite growing interest in attention architectures, convolutional architectures show better robust generalization in the considered setups. \circled{3} Hybrid architectures are a promising avenue in robust fine-tuning. \circled{4} With enough compute, supervised pre-training yields best robust generalization, but multi-modal pre-training is also promising. \circled{5} Robust pre-training is the clear winner in resource constrained fine-tuning settings. \circled{6} When fine-tuning robust backbones with enough compute, using a loss different from the one used for pre-training can boost performance.  \circled{7} Robust pre-training yields limited returns when scaled to larger architectures. \circled{8} Full finetuning is the best overall, and there exist a cost-effective proxy to guide practitioners in finding successful design choices faster.

\paragraph{Scope of the study}  This paper is intended for practitioners aiming to deploy robust vision architectures in the real world under a low training data regime constraint. Architectures beyond vision modality (e.g. language models) need tailored design and fine‑tuning decisions that, in themselves, warrant a dedicated study. Within the vision modality, we consider architectures ranging from 5m to 90m parameters. While we acknowledge that larger architectures exist, architectures in the range considered are adequate to tackle small volumes of training data.
This benchmark focuses on robust classification, while a related computer vision fine-tuning benchmark \citep{goldblum2024battle} focuses on classification and four other computer vision tasks. The main difference is that robust fine-tuning comes with an extra $O(K)$ training compute overhead caused by the necessity to generate $K$ synthetic perturbations for each observation during training. In practice, $K$ is set to 10, which makes robust fine-tuning for classification $10$x more expensive. This computational overhead, as well as the breadth of the configurations considered, is the main reason why this study solely focuses on robust classification.

\paragraph{Related benchmarks}
\cite{tang2021robustart}, \cite{liu2023exploring}, and \cite{shao2021adversarial} benchmark the performance of different 
%categories of 
architectures and training strategies on robustness. A main difference is that they all consider ``training from scratch'' (i.e., training from random initializations). In contrast, our study focuses on fine-tuning from pretrained backbones. Training dynamics observed in one setting do not necessarily transfer to the other 
%, as evidenced in 
\citep{kornblith2019better}. 
% Another key difference is that the current benchmark analyzes configurations with optimized hyper-parameters (see details in Appendix \ref{sec:additional_technical}), while prior works consider fixed hyper-parameters.
The proposed benchmark analyzes configurations with optimized hyper-parameters (see details in Appendix \ref{sec:additional_technical}). This represents an improvement over prior benchmarks on adversarial robustness that consider a single and unique hyper-parameter combination for all the configurations of the benchmark \citep{tang2021robustart,liu2023exploring,shao2021adversarial}.
This study is therefore better geared towards practitioners.
More broadly, this work is inspired by design choices studies in non-robust computer vision \citep{goldblum2024battle} and in robust vision-language \citep{bhagwatkar2024improving}.
% focuses on the impact of design choices on robust generalization, inspired by \citet{goldblum2024battle} in non-robust computer vision and \citet{bhagwatkar2024improving} in robust vision-language.

\section{Design choices}

\begin{wraptable}[12]{R}{0.52\textwidth} 
\centering
\vspace{-10pt}         
\scriptsize
\setlength{\tabcolsep}{3pt}
\renewcommand{\arraystretch}{1.2}
\begin{tabular}{|c|c|c|p{3.5cm}|}
\hline
\textbf{Size} & \textbf{Param. Range} & \textbf{Type} & \textbf{Architectures}\\\hline
\multirow{3}{*}{Small} & \multirow{3}{*}{5–10M} & Conv. & Regnetx004, Efficientnet-b0, Edgenext (small)\\\cline{3-4}
                       &                         & Attn. & DeiT (tiny)\\\cline{3-4}
                       &                         & Hybrid & Coat (tiny), MobileViT (small)\\\hline
\multirow{3}{*}{Medium}& \multirow{3}{*}{25–30M}& Conv. & Convnext (tiny), Resnet50\\\cline{3-4}
                       &                         & Attn. & DeiT (small), ViT (small), Eva02 (tiny), Swin (tiny)\\\cline{3-4}
                       &                         & Hybrid & Coatnet-0\\\hline
\multirow{3}{*}{Large} & \multirow{3}{*}{80–90M}& Conv. & Convnext (base)\\\cline{3-4}
                       &                         & Attn. & ViT (base), Eva02 (base), Swin (base)\\\cline{3-4}
                       &                         & Hybrid & Coatnet-2\\\hline
\end{tabular}
% \vspace{-5pt}          
\caption{Overview of the 19 considered architectures.}
\label{tab:architecture_table}
\end{wraptable}

% \aud{If architecture is a component of the backbone, adjust sentence. Is it 40 backbones x 2 robust loss = 80?}

% We consider $80$ configurations made of $40$ backbones evaluated across $2$ loss objectives. Each configuration is further evaluated across $3$ fine-tuning protocols.
We study $80$ combinations (40 pre-trained backbones $\times$ 2 objective losses) using $3$ fine-tuning protocols over $6$ classification tasks with $C$ classes.
Each observation-label pair $(x,y)$ is drawn i.i.d. from a stationary distribution $(X,Y)$.
Each configuration results in a classifier model $f_\theta:X \rightarrow \Delta_C$, where $\theta$ are the model parameters and $\Delta_C$ denotes the $(C-1)$-dimensional probability simplex. 

% \subsection{Architectures}

% We consider a total of 19 architectures, spanning into three size categories: large (80–90 million parameters), medium (25–30 million), and small (5–10 million). 
% See the summary in Table \ref{tab:architecture_table}. Each architecture is further categorized between one of three structural types: convolutional, attention-based, and hybrid. 

% Small architectures are relevant for deployment in low resource environments (e.g., Jetson Nano, Orion) or with low latency requirements. 
% To our knowledge, this is the first study in robust fine-tuning that considers small size architectures (5-10M) \citep{hua2024initialization,xu2023autolora,hendrycks2019using,liu2023twins}. 
% The largest architectures considered are aligned with existing works \cite{goldblum2024battle, hua2024initialization}. For larger architectures, we refer to works on scaling robustness \cite{wang2024revisiting}.
% \vspace{-5pt}
\subsection{Pre-trained backbones}

Tremendous progress has been made in the development of pre-trained backbones, and each technique is usually followed by multiple variations.
The options available in the open-source community are endless \citep{rw2019timm}, which motivates an extensive benchmarking of pre-trained backbones.

\paragraph{Architectures}

We consider a total of 19 architectures, spanning into three size categories: large (80–90 million parameters), medium (25–30 million), and small (5–10 million) -- see summary in Table \ref{tab:architecture_table}. Each architecture is further categorized between one of three structural types: convolutional, attention-based, and hybrid (i.e., mixture of convolution and attention layers).  Small architectures are relevant for deployment in low resource environments (e.g., Jetson Nano, Orion) or with low latency requirements. To our knowledge, this is the first study in robust fine-tuning that considers small size architectures (5-10M) \citep{hua2024initialization,xu2023autolora,hendrycks2019using,liu2023twins}. The largest architectures considered are aligned with existing works \citep{goldblum2024battle, hua2024initialization}. For larger architectures, we refer to works on scaling robustness \citep{wang2024revisiting}. We acknowledge the potential of large-scale architectures and encourage follow-up work to explore performance at larger scales using our open-source benchmark.

\paragraph{Pre-training protocol}

\begin{wraptable}[12]{R}{0.52\textwidth} 
\centering
\scriptsize
\setlength{\tabcolsep}{3pt}
\vspace{-10pt} 
\renewcommand{\arraystretch}{1.1}
\begin{tabular}{|c|c|p{4cm}|}
\hline
\textbf{Category} & \textbf{Total} & \textbf{Technical Details} \\
\hline
Supervised          & 20 & ImageNet-1k/22k; variants with and without data-aug.\ \& regularization  \\
\hline
Multistep Supervised       & 6  & Imagenet-22k then 1k,  Imagenet-12k then 1k, variants with and without data-aug.\ \& regularization \\
\hline
Robust Supervised          & 5  & 4× APGD-K, 1× PGD-K adversarial pre-training; all based on Classic AT on In1k \\
\hline
Unimodal Self-Sup.       & 4  & MAE, DINO, MIM \\
\hline
Multimodal Self-Sup.      & 3  & CLIP on LAION-2B / LAION-Aesthetics \\
\hline
Fusion          & 2  & CLIP (LAION-2B) followed by fine-pass on Imagenet-1k, and Imagenet-12k/1k \\
\hline
\end{tabular}
\caption{Overview of the 40 considered backbones. }
\label{tab:pre-train_categories}
\end{wraptable}

Prior works have studied the influence of supervised pre-training \citep{hendrycks2019using,mo2022adversarial}, robust pre-training \citep{hua2024initialization, xu2023autolora, liu2023twins}, and multimodal self‑supervised (Multi-SS) pre-training \citep{hua2024initialization} in robust fine-tuning. However, these studies are confined to single architecture types and sizes, which restricts the scope of conclusions. In Section \ref{sec:results}, we will see that some conclusions do no not hold uniformly across all architecture sizes and types.
Furthermore, this study is the first to compare the performance of pre-training protocols such as supervised (multistep), unimodal self-supervised (Uni-SS), and fusion (i.e., mixture of supervised and self-supervised pre-training) in robust fine-tuning. Understanding how such state-of-the-art pre-training protocols contribute to robust generalization remains a knowledge gap for practitioners.
% \vspace{-5pt}
\paragraph{Summary}
% \vspace{-5pt}
Based on the considered architectures and pre-training protocols, a set of $40$ backbones are selected -- see summary in Table \ref{tab:pre-train_categories}. A global summary of the considered backbones, including exhaustive references and Hugging Face identifiers is available in Appendix \ref{sec:configurations}.

\subsection{Fine-tuning protocols} 
\label{sec:fine_tuning}
%\yann{is this classifier always just a linear layer combined with softmax? LP is training of $\theta_2$, so it should be linear. Maybe say: }.

Consider a pre-trained backbone $g_{\theta_1}:X \rightarrow L$, where $L$ denotes an arbitrary latent space.
Further consider a classifier $h_{\theta_2}:L \rightarrow \Delta_{C}$ consisting of a linear layer followed by a softmax.
The goal of fine-tuning is to combine the pre-trained backbone and the classifier together to obtain a final model $f_{\theta}:X \rightarrow \Delta_C$, with $\theta = \{ \theta_1, \theta_2\}$. An observation $x$ is associated to a probability prediction $f_{\theta}(x) = h_{\theta_2} ( g_{\theta_1}(x) )$. The fine-tuning process consists of $E$ epochs over the training dataset.

\paragraph{Full fine-tuning (FFT)} All parameters $\theta=\{ \theta_1, \theta_2\}$ are updated for the downstream task. The proposed FFT setup differs from  prior works~\citep{hua2024initialization, jeddi2020simple}, who employ a single learning rate across the entire model $f_{\theta}$. In contrast, our setup allows for distinct learning rates, $\eta_1$ and $\eta_2$, for $g_{\theta_1}$ and $h_{\theta_2}$, respectively, as well as separate weight decay parameters, $\gamma_1$ and $\gamma_2$.

\paragraph{Linear probing (LP)} Only the classifier layer $h_{\theta_2}$ is updated, while the parameters of the feature extractor are frozen. The learning rate is $\eta_2$ and the weight decay $\gamma_2$. %\yann{do you have weight decay here?}

We consider three fine-tuning protocols: FFT with $E=50$ epochs (denoted FFT-50), FFT with $E=5$ epochs (denoted FFT-5), and LP with $E=50$ epochs (denoted LP-50).
These protocols represent different trade-offs between compute and parameter efficiency. Specifically, LP-50 is parameter-efficient (few trainable weights), while FFT-5 is compute-efficient (short training duration). 
We do not include LP with 5 epochs as it would combine both constraints and would be very restrictive.
Although FFT and LP have been compared before~\citep{hua2024initialization, xu2023autolora, liu2023twins}, there is limited understanding of the compute-efficient setting (FFT-5) and of how design choice combination correlates with performance across the fine-tuning protocols.
 % \aud{settings here = other design components?}

\paragraph{Practical considerations.} 
Our choice of $50$ training epochs is motivated by prior robust fine-tuning works who employ 40 \citep{hua2024initialization} to 60 epochs \citep{xu2023autolora, liu2023twins}.
Other (non-robust) fine-tuning benchmarks have considered more epochs (e.g., $100$ epochs in \cite{goldblum2024battle}) but they are not specifically focused in the low data regime setting. Additional technical details regarding the optimization of hyper-parameters are provided in Appendix \ref{sec:additional_technical}.

%To mitigate overfitting, we rely on the weight decay hyper-parameters and dropout. See Appendix \ref{sec:additional_technical} for more technical details.
%We did not use early stopping because it would induce a non-uniform distribution of training epochs across the considered configurations, which would affect some components of this study (e.g. cross fine-tuning protocols comparisons).
%\yann{This point on overfitting is important, but maybe we could move this to limitations or even wait for reviewers feedback. What do you all think? Is there a risk that overfitting happened and that early stopping could yield very different results? There is not much we can do at this stage.}

% \yann{Here I would expect some information on early stopping if any, and based on what validation loss. If no early stopping was used, we should be prepared to explain why. In particular, given the low-data regime, how do we know that ou models do not overfit with 50 epochs.} \maxi{we did not use early stopping, and no papers in robust fine tuning do, neither 'Battle of the Backbones' paper that does a similar benchmark in CV tasks does; however, we do use other overfitting mitigation techniques like weight decay and dropout. I will discuss this in more details, but let me know if this sufficiently addresses your concern, thanks!}
    
\subsection{Loss objectives}

We consider two loss objectives, namely Classic AT and TRADES \citep{zhang2019theoretically} which are widely popular \citep{wang2023better, croce2020robustbench}. There is no consensus as to which loss to choose to perform robust fine-tuning, as suggested by inconsistent design decisions in the literature (e.g., Classic AT in \cite{hua2024initialization, singh2024revisiting}, TRADES in \cite{xu2023autolora}). 
Although \citet{liu2023exploring} identify TRADES as most effective, their findings are based on models trained from scratch, which differs from fine-tuning where pre-trained backbones play a central role.

\paragraph{Crafting synthetic adversarial perturbations}
Both Classic AT and TRADES require crafting synthetic adversarial perturbations throughout training. 
Given an observation $(x,y)$ and a classifier $f_\theta$, consider the perturbation $x'$ given by the following maximization problem:
\begin{equation} 
\arg\max\limits_{\substack{x' \in \mathbb{B}(x, \epsilon, p)}} \mathcal{L}_{\texttt{CE}} \left( f_\theta(x'), y \right), 
\label{eq:problem}
\end{equation} 
where $\mathcal L_{\texttt{CE} }$ denotes the cross-entropy loss and $\mathbb{B}(x, \epsilon, p) = \{ x \in X : \|x' - x\|_p \leq \epsilon \}$
 is the $\ell_p$-ball around $x$. Projected Gradient Descent (PGD-$K$) \citep{madry2017towards} with $K$ iterations finds an approximate solution $x'_K$ to the perturbation $x'$ resulting from Eq. \ref{eq:problem}. Specifically, PGD-$K$ corresponds to starting from $x'_0=x$ and to iteratively apply the update rule 
\begin{equation} x'_{k+1} = \Pi_{\mathbb{B}(x, \epsilon, p)} \left( x'_{k} + \delta  \texttt{sign}\left(\nabla_{x} \mathcal{L}_{\texttt{CE}} \left( f_\theta(x'_{k}), y \right) \right) \right), \quad {k=0, \ldots, K-1} 
\end{equation}
where $\delta \geq 0$ is the step size and $\Pi_{\mathbb{B}(x, \epsilon, p)}$ is the projection operator to ensure that the perturbed input remains within the $\ell_p$-ball. The APGD-$K$ perturbation \citep{croce2020reliable} improves upon PGD-$K$ by automatically adapting the step size $\delta$, removing the need for manual tuning.

\paragraph{Classic adversarial training (Classic AT)}

Corresponds to training a classifier $f_\theta(\cdot)$ using the cross-entropy loss $\mathcal L_{\texttt{CE}}$ on observations perturbed by APGD-$K$. 
This corresponds to minimizing the loss
$\mathcal{L}_{\texttt{AT}}(x,y):=\mathcal{L}_{\texttt{CE}} (  f_\theta(x'_K), y  )$ over $\theta$. 

\paragraph{TRADES} Corresponds to training a classifier $f_\theta(\cdot)$ with the TRADES loss objective \citep{zhang2019theoretically}. This corresponds to minimizing the following loss over $\theta$: 
\begin{equation}
\mathcal{L}_{\texttt{TRADES}}(x,y) := 
    \mathcal{L}_{\texttt{CE}}(f_\theta(x), y)  + \beta\,\texttt{KL}(f_\theta(x) \| f_\theta(x'_{K})),
\end{equation}
where the scalar $\beta \geq 0$ controls the trade-off between cross-entropy and the Kullback–Leibler (KL) divergence of predictions on perturbed and unperturbed inputs.

\paragraph{Practical considerations} To facilitate comparison between Classic AT and TRADES loss objectives, we always consider the same process to craft synthetic adversarial perturbations, namely APGD-$K$ with $K=10$, $\epsilon=4/255$, and bounded with respect to the $\ell_\infty$-norm. Additionally, the training data is augmented regardless of the loss objective using standard augmentation techniques (see Appendix \ref{sec:additional_technical} for more details).

\section{Threat definition and evaluation methods for robust generalization}

% \yann{I still feel like setting the generalization objective before explaining the fine-tuning objective is more natural. If we keep it in this order, I would start this section with something like:   }

Although we set a specific type of adversarial perturbation for the optimization strategy (i.e., APGD-K for $\ell_\infty$-norm), deploying machine-learning systems reliably and responsibly requires generalization to diverse, unknown and evolving types of perturbations. We now define additional perturbation types that the model will face at test time, to evaluate robust generalization.

\subsection{Threat model at test time} 
\label{sec:threat_models} 

To study robust generalization, we define the \textit{threat model} which specifies the possible perturbation types faced by the model at test-time \citep{akhtar2018threat}.
We use the notation $\mathcal T_X(z)$ to denote the distribution of observations drawn from $X$ that contain a perturbation of type $z$. We consider a finite set of perturbation types noted $\tau$, where each type can be categorized into \textit{adversarial} and \textit{common} perturbations.

The adversarial perturbations are bounded by a scalar $\epsilon$ with respect to the $\ell_p$-norm, i.e. $x \sim X, x^\prime \sim \mathcal T_X(z)$ such that $ \lVert x - x^\prime \rVert_p \leq \epsilon$. We include three adversarial perturbation types, generated from $p=1,2,\infty$, and $\epsilon=75.0, 2.0, 4/255$, respectively. The values for $\epsilon$ are standard choices in robustness benchmarks~\citep{croce2020robustbench, singh2024revisiting}.
For each norm, we employ the AutoAttack framework \citep{croce2020reliable} to generate 4 types of attacks totalling 4 x 3 = 12 types of adversarial attacks. The 4 types of attacks comprise black-box (assume no prior access to the architecture) and white box (assume prior access to the architecture) attacks. 

The common perturbations reflect unfortunate events that commonly occur in vision systems (e.g. noise, blur, contrasts, digital format compressions, etc) and that hamper the predictive performance. 
We employ \citet{imgaug, hendrycks2019benchmarking} to generate common perturbations, generated among 11 types, including weather variations (snow, 2 types of noise, 3 types of blur, contrast, brightness), digital transformations (jpeg compression, pixelation and elastic transformations).

\paragraph{Summary of the threat model}

In summary, the threat model is $\mathcal T_{X}(z), z\in \tau =  \{ \emptyset, \infty, 1, 2, \texttt{common} \}$, where $\tau$ comprises five perturbation types: no perturbations (i.e., clean observations, noted $\emptyset$), adversarial perturbations under the $\ell_1$, $\ell_2$, and $\ell_\infty$ norm (noted $1$, $2$ and $\infty$), and common perturbations (noted $\texttt{common})$. In total, 23 types of perturbations are evaluated at test time.
Appendix \ref{sec:crafting_practice} describes how we generate these test-time perturbations using open-source software AutoAttack~\citep{croce2020reliable} and \citet{imgaug,hendrycks2019benchmarking}.

\paragraph{Considerations on the synthetic-to-real-world robustness gap} In this study, we evaluate robustness using adversarial and common perturbations that are synthetically generated. Prior works have shown that in the training-from-scratch regime, robustness to synthetic common perturbations is only weakly correlated with robustness to natural distribution shifts \citep{taori2020measuring}. However, subsequent work \citep{laugros2021using} demonstrates that this correlation strengthens significantly when a broad range of synthetic perturbations is considered. Our benchmark includes 23 perturbation types across 6 datasets of varying difficulty, supporting this broader evaluation. Although a gap with the real world remains, we emphasize that the primary goal of the study is to better understand which design choices contribute to improving robustness to perturbations unseen during training. Synthetic perturbations enable the design of a systematic, scalable, and reproducible experiment to understand this problem \citep{hendrycks2019benchmarking}.

\subsection{Evaluating robust generalization} 

We measure performance against the threat model using the \textit{accuracy}, which corresponds to the total number of correct predictions over the total number of observations in the test dataset. 
Accuracy is the standard evaluation metric in the field of robust classification. For instance, state-of-the-art improvements are typically reported using accuracy \citep{croce2020robustbench}. Prior works on robust fine-tuning \citep{tang2021robustart,liu2023exploring,shao2021adversarial} and more broadly on robustness \citep{pmlr-v202-wang23ad}, predominantly focus on accuracy. We therefore adopt accuracy as our primary metric to ensure consistency with existing literature and to facilitate comparisons.

Recall that a configuration is the combination of a pretrained backbone and a loss objective that results into a classifier model. For each of the three fine-tuning protocols considered (FFT-50, FFT-5 and LP-50), we evaluate the performance of each configuration as follows. 
For every configuration $i \in \{1, \dots, I\}$ on dataset $d\in \{1, \dots, D\}$ we obtain a predictive accuracy score $a_{i,d}(z)\in [0,1]$ for each perturbation type $z \in \tau$. Let $\mathbf{a}_{i,d} := \left[ a_{i,d}(z_1), \dots, a_{i,d}(z_{|\tau|})\right]$ denote the vector of predictive accuracies of configuration $i$ on dataset $d$. 

\paragraph{Borda score} We use the Borda score to compare the relative performance of various configurations on the same fine-tuning protocol.
Consider any pair $v=(d,z)$, consisting of a dataset $d$ and a perturbation type $z$ as a voter. Let $V= D\times|\tau|$ be the set of all voters. 
To each voter $v=(d,z)$ corresponds a function $\texttt{rank}_v:I\to \{1,\ldots, |m_v|\}$ that ranks the configurations $i\in I$ based on their score $a_{i,d}(z)$, in decreasing order. 
The configuration $i_\text{top}$ with best performance gets rank 1 (i.e., $\texttt{rank}_v(i_\text{top})=1$) and the worst one gets rank $m_v$.
We have $m_v\leq |I|$ to account for the possibility of equal scores (and so equal ranks). 
Then, the Borda score for each configuration $i\in I$ is defined by $B(i) := \sum_{v = 1}^V  m_v-\texttt{rank}_v(i)$. 

\paragraph{Sum score} To account for absolute performance and to compare configurations across different fine-tuning protocols we use the \textit{Sum score}.
For each configuration $i\in I$ the sum score is defined by $S(i):=\sum_{(d,z)\in V}a_{i,d}(z)$. By summing the accuracy scores across all perturbation types and datasets, the sum score rewards peak performance even when the accuracy is inconsistent. This contrasts with the Borda score that penalizes inconsistent performance through ranking degradation. 

\paragraph{Mean Absolute Correlation} For each dataset $d$, we define a $|\tau| \times |\tau|$ Spearman correlation matrix $\mathbf{C}^{(J, d)}$ computed over the accuracy vectors $\{ \mathbf{a}_{i,d} \}_{j=1}^J$ associated to a subset of configurations $J \subseteq I$. 
The subset $J$ can represent all the configurations ($J=I$) or a subset with a common specific characteristic (e.g. architecture type, etc).
The \textit{mean absolute correlation} for dataset $d$ on the subset of configurations $J$, is noted $\texttt{MAC}^{(d,J)}$, and is given by
$ \frac{1}{|\tau|(|\tau|-1)} \sum_{l \neq k} |C_{l,k}^{(J,d)}| $.
The $\texttt{MAC}^{(d,J)}$ is the average absolute off-diagonal correlation between all pairs of perturbation types on dataset $d$ for the subset of configurations in $J$.
A high $\texttt{MAC}^{(d,J)}$, close to $1$, indicates that, on average, the performance across all perturbation types is strongly related, suggesting more consistent or uniform robust generalization. Lower $\texttt{MAC}^{(d,J)}$ values indicate that on average the performance across all perturbation types is less correlated, implying that robustness may be specific to certain perturbations rather than uniform.
We also compute a global average across datasets: $\texttt{MAC}^{(\texttt{avg}, J)} = \frac{1}{D} \sum_{d=1}^D \texttt{MAC}^{(d,J)}$. This informs us on the strength of the robust generalization pattern across datasets.

\paragraph{Summary on the ranking methodology} The methodology to obtain rankings relies on three distinct aggregation methodologies (total sum score, the Borda score and the Mean Absolute Correlation score) to avoid biasing the conclusions to a specific score. The rankings for each score are available in the Appendix (Tables \ref{tab:global_ranking_FFT50epochs}, \ref{tab:global_ranking_FFT5epochs} and \ref{tab:global_ranking_LP50epochs}). Researchers interested in granular performance (with no aggregation) can refer to the Github of the project where the detailed measurements are available for further analysis

% \yann{From what I understand $\texttt{MAC}_d$ is defined for every dataset. The last sentence does not make sence to me. A value $C^{d}_{i,j}$ that is high, namely close to $1$, means that, for dataset $d$ and threat model $i$, performance correlates with that of threat model $j$ over the set of configurations.  In particular, configurations with best performance for one threat model also tend to be good configurations with respect to every other threat model. Similarly, averaging over datasets } 
%For a dataset $d \in {1, \dots, D}$, 

\section{Results}
\label{sec:results}

We select $6$ datasets that fit in the \textit{low data regime} (details in Appendix \ref{sec:additional_technical}). 
We consider five datasets from the natural image domain (\textbf{Caltech101} \citep{1384978}, \textbf{Aircraft} \citep{maji2013fine}, \textbf{Flowers} \citep{nilsback2008automated}, \textbf{Oxford pet} \citep{parkhi2012cats}, \textbf{Stanford cars} \citep{krause20133d} ) and one from the satellite imagery domain (\textbf{Land-Use} \citep{yang2010bag}). 

\begin{wrapfigure}[14]{r}{0.48\textwidth}
  % \vspace{-10pt}  % pull figure up slightly if needed
  \centering
  \includegraphics[width=\linewidth]{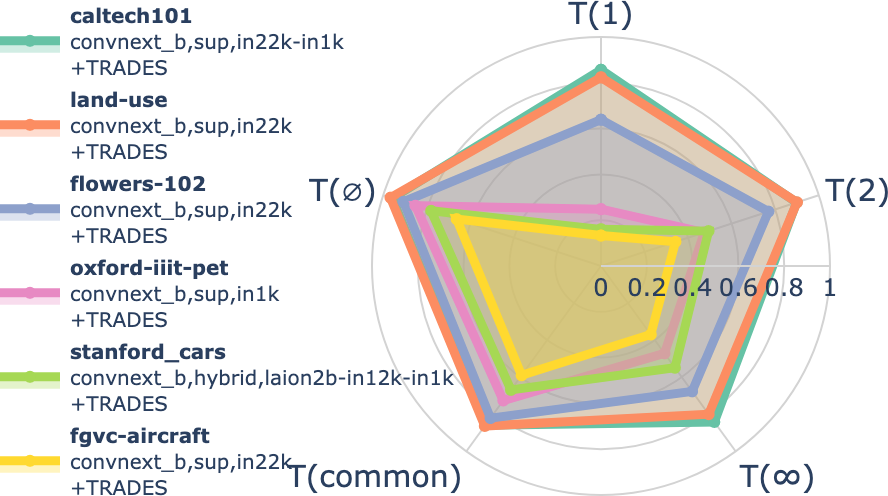}
  \captionsetup{format=plain}  % ensures caption doesn't push figure down
  \caption{\small Robust generalization of the best configuration per dataset using FFT-50 (Borda score).}
  \label{fig:best_per_dataset}
  % \vspace{-6pt}
\end{wrapfigure}

We benchmark $240$ design combinations ($40$ pre-trained backbones $\times$ $2$ robust losses $\times$ $3$ fine-tuning protocols) over $6$ datasets, totaling $1,440$ evaluated configurations. 

The hyper-parameters of each configuration are independently optimized (details are reported in Appendix \ref{sec:additional_technical}). 
Each configuration is tested against $5$ perturbation types, unseen during (pre-) training (Section~\ref{sec:threat_models}), resulting in $7,200$ robustness measurements.
To our knowledge, this benchmark includes the most diverse and comprehensive set of design choices in the robust fine-tuning setting. 
% Collected measurements, and code are open-sourced.
The code base \footnote{\url{https://github.com/MaxHeuillet/robust_training}} open sources the optimal hyper-parameter combinations found for each fine-tuning configuration, on each of the 6 datasets. These can be used by researchers as a starting point to their own research, reducing redundant compute.

\subsection{Which configurations perform best?}

%\aud{What is the take home message of Table~\ref{tab:FFT_50_epochs}?} 
%\maxi{Mhhm, i feel this table is not for analysis, it is more the global winners, split between architecture size, for practitioners to pick. How do you feel abt this as take home message? }

\paragraph{Best performing configurations overall.} Table~\ref{tab:FFT_50_epochs} reports the best performing configurations in FFT-50. We see that the best performing backbone is convolutional (\textit{Convnext (base), with supervised pretraining on Imagenet-22k, using TRADES}). 
FFT-50 clearly outperforms other fine-tuning protocols, with a best sum score of $19.79$, which is $61\%$ higher than FFT-5 and $53\%$ higher than LP-50 (see Table \ref{tab:FFT_5_epochs} and \ref{tab:LP_50_epochs} in Appendix). In the Appendix, Tables \ref{tab:global_ranking_FFT50epochs}, \ref{tab:global_ranking_FFT5epochs}, and \ref{tab:global_ranking_LP50epochs} report the ranking of all the configurations across FFT-50, FFT-5, and LP-50 respectively.

\paragraph{Best performing configurations per dataset.}
Figure~\ref{fig:best_per_dataset} displays the best performing configuration per dataset, when using FFT-50. 
We observe that convolutional architectures outperform other options on all considered datasets. 
Despite growing attention on the robustness of attention-based architectures \citep{bai2021transformers,liu2023exploring,shao2021adversarial}, our findings show that the robust generalization capacity of well tuned convolutional architectures should not be underestimated.
Additionally, on two datasets (Caltech101, and Land-Use), the best configurations achieve accuracy above $0.8$ on all perturbation types, demonstrating strong robust generalization. This performance is remarkably high for the field \citep{croce2020robustbench}, demonstrating the practical potential of robust fine-tuning and the importance of carefully identifying best design choices.
%Specifically, Land-Use belongs to the satellite imagery domain, a domain where robustness is critical.
%This suggests that low data regimes might be better served by simpler designs.

\paragraph{Low-cost proxies exist in robust fine-tuning.}
Given the evolving set of available design choices, practitioners need to be equipped with low-cost tools to rapidly identify design choices that are more promising than others. 
The identification of low-cost proxies helped practitioners in natural language modeling \citep{zhu2022predicting}, and can also benefit costly protocols such as robust learning.
Between LP-50 and FFT-5, we find that LP-50 is the most reliable low-cost proxy to FFT-50 (see Figure \ref{fig:FFT50_LP50}), especially when using TRADES over Classic AT. Indeed, the correlation between LP-50 and FFT-50 using TRADES is the highest.

% \begin{tcolorbox}[mysummarybox]
% \textbf{Summary 1:} FFT-50 emerges as the best overall setup for robust fine-tuning and its most reliable cost effective proxy is LP-50 with TRADES. Additionally, convolutional architectures with supervised pre-training consistently outperform others in low-data regimes. 
% \end{tcolorbox}

\begin{table}[ht]
\centering
\resizebox{\textwidth}{!}{%
\begin{tabular}{|c|c|c|c|}
\toprule
\textbf{Size} &                                                                              \textbf{Gold (1st)} &                                                                          \textbf{Silver (2nd)} &                                                                          \textbf{Bronze (3rd)} \\
\midrule
        small &           \makecell{coat\_t,sup,in1k, TRADES \\[0.3ex] \footnotesize (GR:18, BS:1653, SS:15.83)} &     \makecell{edgenetx\_s,sup,in1k, TRADES \\[0.3ex] \footnotesize (GR:23, BS:1552, SS:14.66)} & \makecell{edgenetx\_s,sup,in1k, Classic AT \\[0.3ex] \footnotesize (GR:33, BS:1356, SS:12.88)} \\
       medium & \makecell{convnext\_t,sup,in22k-in1k, TRADES \\[0.3ex] \footnotesize (GR:14, BS:1773, SS:16.49)} &      \makecell{convnext\_t,sup,in1k, TRADES \\[0.3ex] \footnotesize (GR:15, BS:1681, SS:15.6)} &    \makecell{convnext\_t,sup,in22k, TRADES \\[0.3ex] \footnotesize (GR:20, BS:1650, SS:15.07)} \\
        large &       \makecell{convnext\_b,sup,in22k, TRADES \\[0.3ex] \footnotesize (GR:1, BS:2281, SS:19.79)} & \makecell{coatnet\_2,sup,in12k-in1k, TRADES \\[0.3ex] \footnotesize (GR:2, BS:2127, SS:18.74)} &      \makecell{coatnet\_2,sup,in12k, TRADES \\[0.3ex] \footnotesize (GR:3, BS:2116, SS:18.87)} \\
\bottomrule
\end{tabular}
}
\vspace{0.5em}
\caption{Top FFT-50 configurations, with global ranking (GR) based on Borda score (BS), sum score (SS) also reported below. }
\label{tab:FFT_50_epochs}
\end{table}

 % \yann{Sum Score? it does not seem to be defined anywhere?}

\subsection{Design Choices Favoring TRADES in Robust Fine-Tuning.}

\paragraph{Overall, TRADES outperforms Classic AT.}
It has been shown previously that TRADES outperforms Classic AT when training from scratch \citep{liu2023exploring}.
Our results show that these conclusions hold in the fine tuning setting as well (see Figure \ref{fig:main_effects} in the Appendix). 
We next extend these results by identifying strong interactions between the loss and other design choices in the FFT-50 setting (see Figure~\ref{fig:interactions}). The identification of such interactions with TRADES is particularly valuable, given its frequent association with state-of-the-art performance on robustness benchmarks \citep{wang2023better, croce2020robustbench}.

\paragraph{TRADES interacts positively with architecture size.}
On average, TRADES achieves higher returns compared to Classic AT when architecture size grows (see Figure~\ref{fig:sub1}). 
Additionally, the odds ratio of TRADES outperforming Classic AT increases steeply with architecture scale, which is a significant effect in FFT-50 (see Figure \ref{fig:TRADES_odds} in Appendix \ref{sec:additional_results}). These results suggest that TRADES is a promising approach to improve the robustness of large systems, a setting where Classic AT is currently the preferred approach~\citep{wang2024revisiting}. 
Existing implementations of TRADES require the storage of two forward passes in memory, which motivates an avenue to improve this algorithmic limitation to fully reveal the potential of TRADES on large architectures. 

\paragraph{TRADES interacts best with convolutional and hybrid architectures.}
While TRADES and Classic AT yield equivalent outcomes (similar mean Borda score) for attention-based architectures, convolutional and hybrid architectures benefit most from using TRADES over Classic AT (see Figure \ref{fig:sub2}). Since convolutional architectures tend to overfit more local features and patterns \citep{bhojanapalli2021understanding}, this result suggests that TRADES regularizes convolutional architectures more efficiently than Classic AT in computer vision tasks.

\subsection{Distinct robust generalization patterns across architectures sizes and types.} 

\begin{figure}[H]
    \centering
    \begin{subfigure}[t]{0.3\textwidth}
        \includegraphics[width=\linewidth]{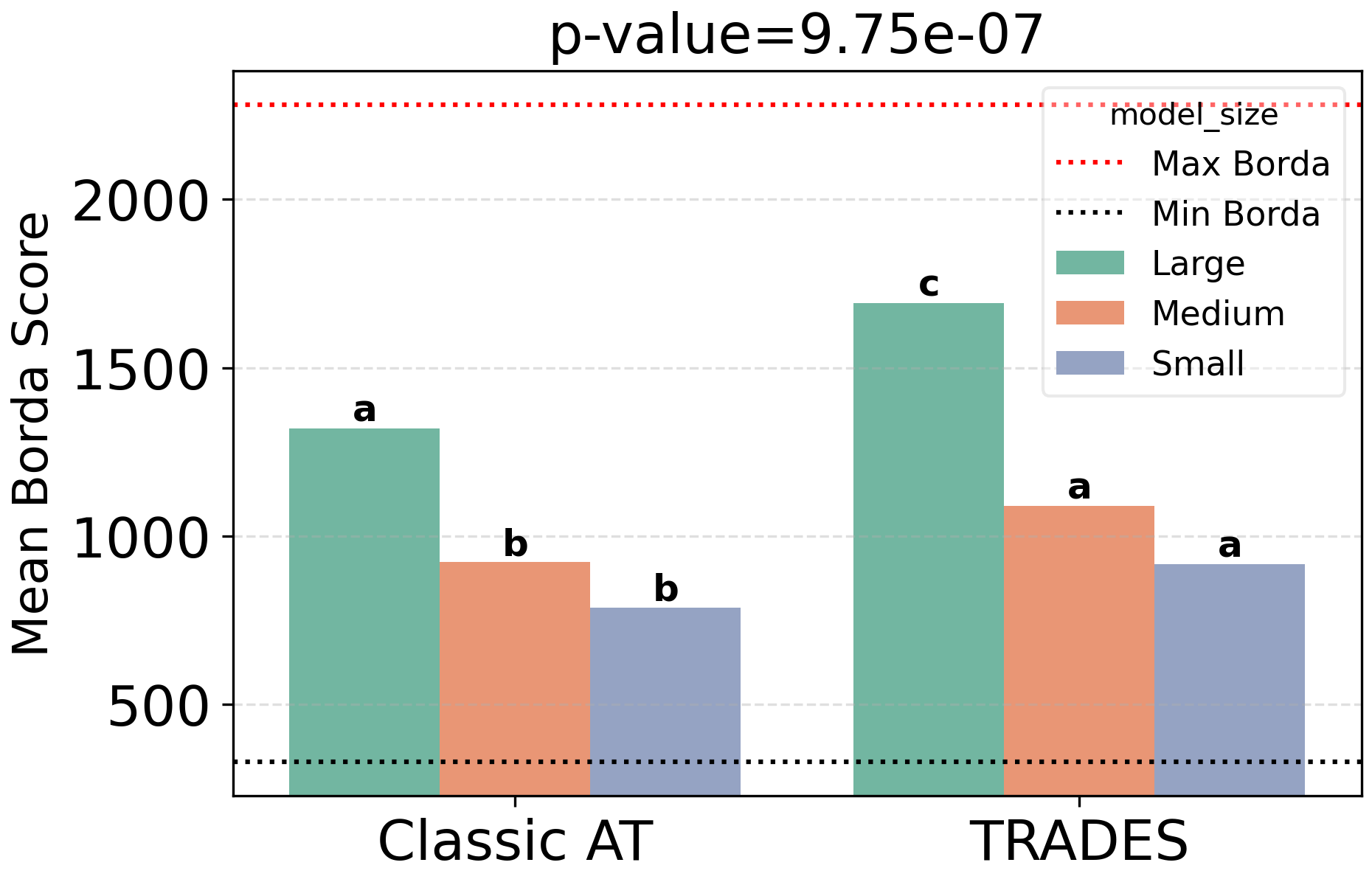}
        \caption{Loss Function $\times$ Arch. Size}
        \label{fig:sub1}
    \end{subfigure}
    \hfill
    \begin{subfigure}[t]{0.3\textwidth}
        \includegraphics[width=\linewidth]{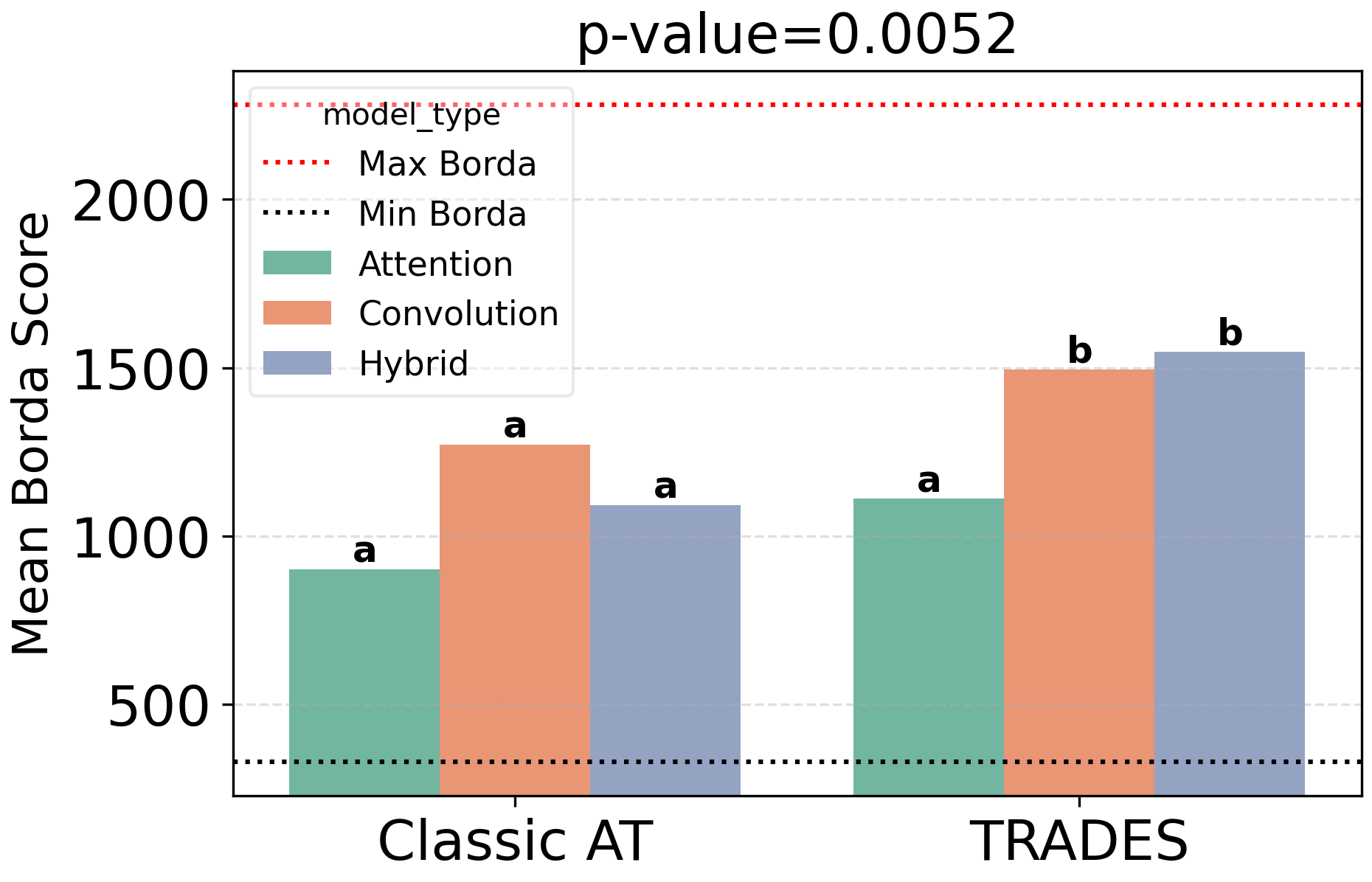}
        \caption{Loss Function $\times$ Arch. type}
        \label{fig:sub2}
    \end{subfigure}
    \hfill
    \begin{subfigure}[t]{0.3\textwidth}
        \includegraphics[width=\linewidth]{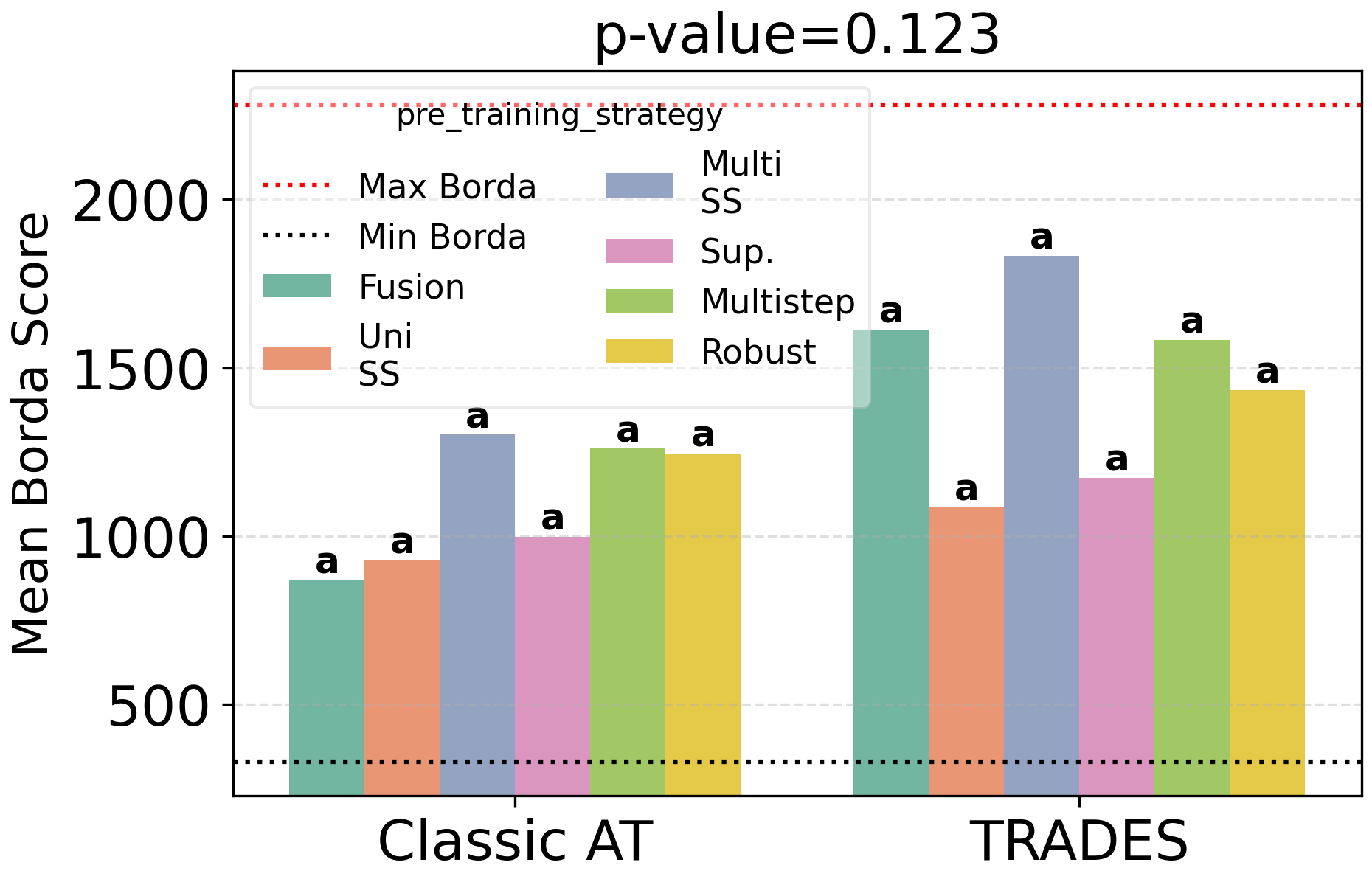}
        \caption{Loss Function $\times$ Pre-training}
        \label{fig:sub3}
    \end{subfigure}

    \vspace{0.5cm}
    \begin{subfigure}[t]{0.3\textwidth}
        \includegraphics[width=\linewidth]{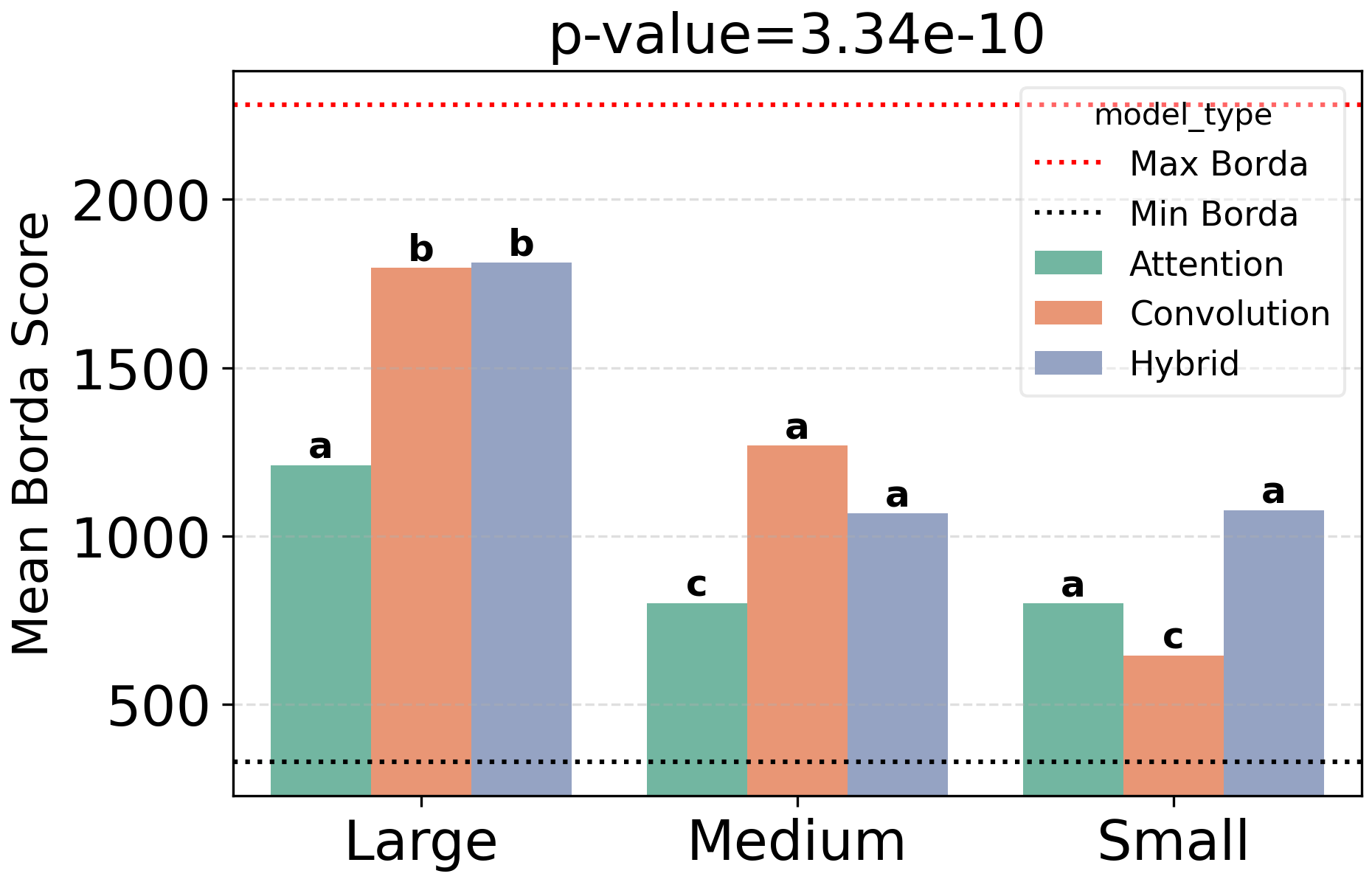}
        \caption{Arch. Type $\times$ Size}
        \label{fig:sub4}
    \end{subfigure}
    \hfill
    \begin{subfigure}[t]{0.3\textwidth}
        \includegraphics[width=\linewidth]{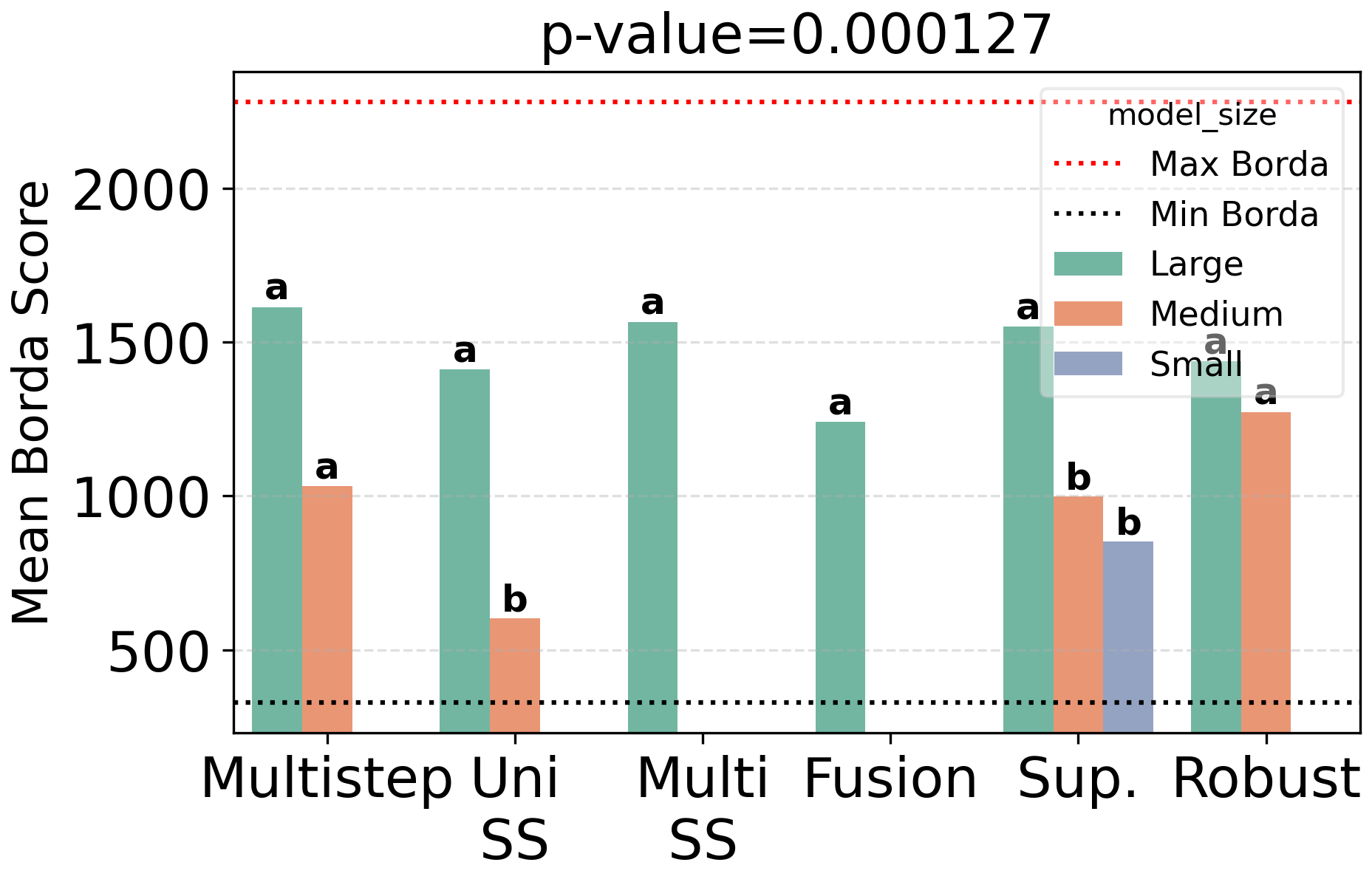}
        \caption{Pre-training $\times$ Arch. Size}
        \label{fig:sub5}
    \end{subfigure}
    \hfill
    \begin{subfigure}[t]{0.3\textwidth}
        \includegraphics[width=\linewidth]{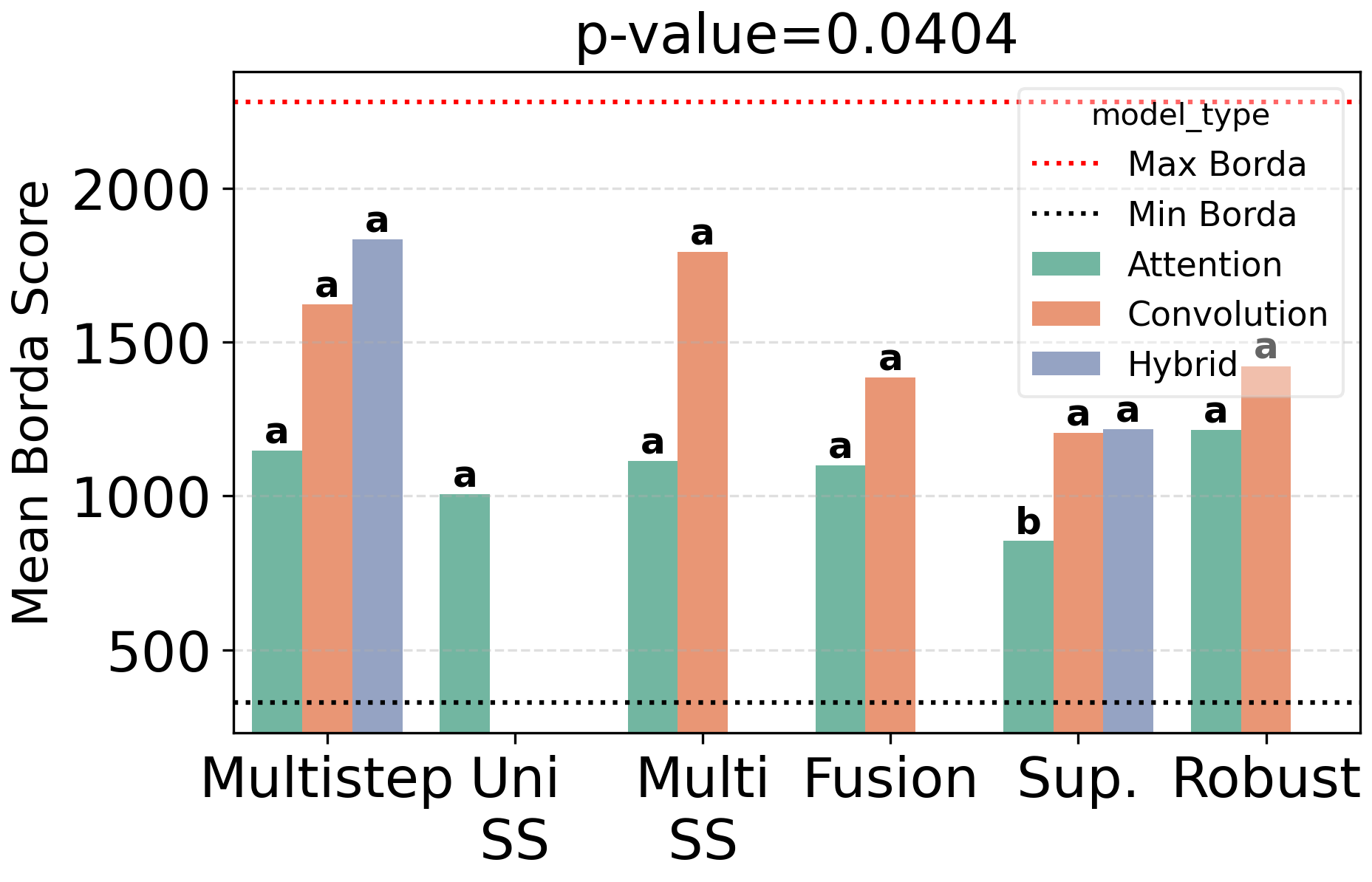}
        \caption{Pre-training $\times$ Arc. Type}
        \label{fig:sub6}
    \end{subfigure}

    \caption{Nested Welch's ANOVA of the form A $\times$ B testing the main effect of A and how the effect of B varies within each level of A (p-value on top).  Post-hoc groupings from Tukey HSD tests are annotated with letters above the bars: bars with different letters belong to significantly different groups, at the $90\%$ confidence level. Results for FFT-50. }
    \label{fig:interactions}
\end{figure}

\paragraph{Larger architectures are consistently better.}
Large architectures clearly outperform medium and small architectures in FFT-50 (see Table \ref{tab:FFT_50_epochs}) and generalize better (see Table \ref{fig:correlation_summary}).
Large convolutional and hybrid architectures outperform attention-based ones on average (see Figure~\ref{fig:sub4}), though attention models may show their full potential at larger scales \citep{wang2024revisiting}.
Because model scale is often subject to limitations in practice, we also provide analysis at specific architecture sizes to guide practitioners with such limitations.

\paragraph{If constrained to small architectures, hybrid architectures are the best option.}
Among small architectures, hybrid architectures achieve significantly higher scores than fully convolutional ones (see Figure \ref{fig:sub4}). Using TRADES loss, \textit{Coat (tiny)} and \textit{EdgeNetx} rival larger architectures and achieve impressive rankings of GR:18 and GR:23, corresponding to tier-1 performance among $80$ configurations (see Table \ref{tab:FFT_50_epochs}).
Prior works on robustness are generally focused on larger architectures \citep{liu2023exploring}, but this result extends knowledge by highlighting the practical potential of hybrid architectures for robust fine-tuning using small architectures on data scarce regimes. The result also constitutes valuable motivations for the community that supports hybrid architectures \citep{dai2021coatnet,Maaz2022EdgeNeXt,dai2021coatnet}. 

\paragraph{Promising generalization properties of hybrid architectures.} 
Table \ref{tab:spearman_corr} shows that hybrid architectures achieve the highest MAC values compared to attention and convolutional architectures in FFT-50. This finding complements prior work demonstrating strong robustness of hybrid architectures trained from scratch \citep{liu2023exploring}, extending these conclusions to the robust fine-tuning setting.
% \citet{liu2023exploring} point out a strong performance for the hybrid architecture, but this observation was for the random initialization setting for one specific hybrid architecture. Our results complement this observation because we show that hybrid architectures also perform well in the robust fine-tuning setting. 
We also generalize the observation across diverse scales and types of hybrid architectures, while the previous observation held only for CoatNet (16M). Finally, Table \ref{tab:spearman_corr}  provides a precise characterization on the robust generalization capability of hybrid architectures, beyond accuracy score.
Modern learning frameworks distinguish pre-training from post-training phases. Pre-training aims to develop general-purpose representations, while post-training focuses on fine-tuning these representations to specific tasks or preferences. This workflow has shown successful returns in other fields of machine learning \citep{devlin2019bert}. Within this context, the problem of \textit{robust pre-training} seeks to simultaneously achieve general-purpose representation and specific alignment to robustness during pre-training, which may explain the lower returns.

%our \julien{il manque pt un mot après our}

% \begin{tcolorbox}[mysummarybox]
% \textbf{Key Takeaway 3:} 
% \end{tcolorbox}

\subsection{Influence of the pre-training strategy on robust generalization}

% \maxi{Introductory sentence?}

\paragraph{Multi-modal self-supervised pretraining is beneficial for convolutional architectures.} 
In FFT-50, the \textit{ Convnext (base) } architecture with multimodal self-supervised pre-training using the TRADES loss achieves the fourth best ranking in terms of Borda score, and top-2 in terms of sum score (see Table \ref{tab:global_ranking_FFT50epochs} in Appendix \ref{sec:additional_results}). 
This performance is surprising given prior results \citep{hua2024initialization} showing that supervised and robust pre-training often outperform Multi-SS. 
However, prior works focus on Multi-SS of attention-based architectures, while the reported performance improvement targets Multi-SS on convolutional architectures (see Figure \ref{fig:sub5}). The potential of Multi-SS pre-training is further evidenced in Figure \ref{fig:sub3}, where Fusion (which is based on Multi-SS) pre-training achieves second best average performance, behind Multi-SS. 

% \paragraph{Multistep and Uni-SS pre-trainings may offer local performance gains, but are less competitive globally.}
% \maxi{By local performance gain i mean that for example, i observe that top performing attention architectures benefit from multistep, but yet they dont make it to the top global ranking.}
% \maxi{I don't succeed yet in getting this discussion point, in the sections above we claim novelty on the study of SOTA pre-training like self-supervised (unimodal), and multi-step supervised, but the patterns are difficult to interpret. I don't know yet how to get well with this discussion yet. } Under FFT-50, the second best approach using Borda score was obtained with mutli-step pre-training, but the performance gap with the same architecture with supervised pre-training is marginal (less than $1\%$ sum and Borda score variation).  On average, multi-step pretraining achieves the 2nd best score after multi-modal pre-training. Figure \ref{fig:sub6}. 

\paragraph{Robust pre-training performs best in constrained fine-tuning protocols.} 
In FFT-5, the global gold (global rank GR:1), silver (GR:2), and bronze (GR:3) are achieved with robust pretraining (see Table \ref{tab:FFT_5_epochs} in Appendix \ref{sec:additional_results}). Similarly, the top-3 in LP-50 (GR:1,2 and 3) are also achieved with robust pre-training (see Table \ref{tab:LP_50_epochs} in Appendix \ref{sec:additional_results}). 
Note that this competitive performance does not hold in the less constrained FFT-50 protocol, where the best configuration based on robust pre-training achieves a global ranking of 9 (see Table \ref{tab:global_ranking_FFT50epochs}). 
Our results are aligned with prior results showing that robust pre-training helps in parameter-efficient settings such as low-rank adaptation \citep{xu2023autolora, liu2023twins} and linear probing \citep{hua2024initialization}. We further show that robust pre-training is also beneficial in fine-tuning protocols constrained on the number of updates, a setting not covered by prior works.
% Prior works confirm that robust pre-training helps in parameter efficient settings such as low-rank adaptation \citep{xu2023autolora, liu2023twins} and linear probing \citep{hua2024initialization}. Our results confirm this observation, and further show that robust pre-training is also beneficial in fine-tuning protocols constrained on the number of updates, a setting not covered by prior works.

\paragraph{Robust pre-training of larger architectures (may) have limited returns.}
Given its high computational cost, robust pre-training prompts critical evaluation of its return on investment relative to alternative pre-training protocols.
The performance gains from medium to large architectures are relatively modest for robust pre-training (see Figure \ref{fig:sub5}).
The rate of improvement \footnote{Rate measured using the relative change w.r.t. the average of the two scores, and computed as
$\frac{\text{large} - \text{medium}}{(\text{large} + \text{medium}) / 2}$. This rate to ensures a symmetric and unbiased comparison that does not privilege either model scale.} is $+12\%$ for robust pre-training, whereas it reaches $+43\%$ for supervised, $+44\%$ for multi-step, and $+80\%$ for Uni-SS in FFT-50.
This relatively low gain is due to already high performance of robust pre-training at the medium scale, leaving less room for improvement at the larger scale. 
Although recent works have emphasized scaling robust pretraining to large architectures \cite{singh2024revisiting}, there are currently no robust pre-trained small architectures. This finding suggests that robust pre-training at smaller architecture scales could be a promising and underexplored direction for future research.

\paragraph{Influence of loss objective switches between robust pre-training and robust fine-tuning.} 
When specifically considering robust pre-trained architectures, a question for practitioners is: \textit{should we use the same robust loss objective for fine-tuning as for pre-training?}
With FFT-50, configurations that use a different loss objective for robust fine-tuning than for robust pre-training significantly outperform configurations that use the same loss for both phases. 
While the global mean between both choices are not statistically different (e.g., $p=0.42$ with the Mann–Whitney test), we observe that switched configurations are strongly overrepresented among top performers across the $6$ datasets considered: 5 out of 6 top-1 configurations use a loss switch, with a binomial $p$-value of $0.041 < 0.05$. Previous works have used switching \citep{xu2023autolora, liu2023twins} and non-switching strategies \citep{hua2024initialization}. Our finding provides the first evidence that switching losses between pre-training and fine-tuning can be beneficial with enough compute (result holds only in FFT-50).

\begin{table}[h]
\resizebox{\textwidth}{!}{%
\begin{tabular}{ll|r|r|r|r|r|r|r|}
\cline{3-9}
                                                                                                                            &                               & Caltech101                   & Aircraft                     & Flowers-102                  & Oxford-pet                   & Stanford-cars                & Land-use                     & Global $\texttt{MAC}_{\texttt{avg} }$              \\ \hline
\multicolumn{2}{|l|}{\begin{tabular}[c]{@{}l@{}}MAC per dataset $\texttt{MAC}_{d}$ over \\ the $80$ configs.\end{tabular}}                                               & 0.847                        & 0.782                        & 0.805                        & 0.681                        & 0.849                        & 0.807                        & 0.795                        \\ \hline
\rowcolor[HTML]{C0C0C0} 
\multicolumn{1}{|l|}{\cellcolor[HTML]{C0C0C0}}                                                                              & Classic AT                    & 0.876                        & 0.702                        & 0.842                        & 0.778                        & 0.844                        & 0.890                        & 0.822                        \\ \cline{2-9} 
\rowcolor[HTML]{C0C0C0} 
\multicolumn{1}{|l|}{\multirow{-2}{*}{\cellcolor[HTML]{C0C0C0}\begin{tabular}[c]{@{}l@{}}Loss\\ objective\end{tabular}}}    & TRADES                        & 0.823                        & 0.884                        & 0.830                        & 0.669                        & 0.896                        & 0.777                        & 0.813                        \\ \hline
\multicolumn{1}{|l|}{}                                                                                                      & {\color[HTML]{000000} Large}  & {\color[HTML]{000000} 0.825} & {\color[HTML]{000000} 0.882} & {\color[HTML]{000000} 0.860} & {\color[HTML]{000000} 0.859} & {\color[HTML]{000000} 0.871} & {\color[HTML]{000000} 0.803} & {\color[HTML]{000000} 0.850} \\ \cline{2-9} 
\multicolumn{1}{|l|}{}                                                                                                      & {\color[HTML]{000000} Medium} & {\color[HTML]{000000} 0.743} & {\color[HTML]{000000} 0.628} & {\color[HTML]{000000} 0.743} & {\color[HTML]{000000} 0.436} & {\color[HTML]{000000} 0.701} & {\color[HTML]{000000} 0.833} & {\color[HTML]{000000} 0.681} \\ \cline{2-9} 
\multicolumn{1}{|l|}{\multirow{-3}{*}{\begin{tabular}[c]{@{}l@{}}Architecture\\ size\end{tabular}}}                         & {\color[HTML]{000000} Small}  & {\color[HTML]{000000} 0.911} & {\color[HTML]{000000} 0.531} & {\color[HTML]{000000} 0.503} & {\color[HTML]{000000} 0.467} & {\color[HTML]{000000} 0.575} & {\color[HTML]{000000} 0.678} & {\color[HTML]{000000} 0.611} \\ \hline
\rowcolor[HTML]{C0C0C0} 
\multicolumn{1}{|l|}{\cellcolor[HTML]{C0C0C0}}                                                                              & Attention                     & 0.884                        & 0.800                        & 0.761                        & 0.713                        & 0.856                        & 0.882                        & 0.816                        \\ \cline{2-9} 
\rowcolor[HTML]{C0C0C0} 
\multicolumn{1}{|l|}{\cellcolor[HTML]{C0C0C0}}                                                                              & Convolutional                 & 0.762                        & 0.756                        & 0.811                        & 0.574                        & 0.818                        & 0.719                        & 0.740                        \\ \cline{2-9} 
\rowcolor[HTML]{C0C0C0} 
\multicolumn{1}{|l|}{\multirow{-3}{*}{\cellcolor[HTML]{C0C0C0}\begin{tabular}[c]{@{}l@{}}Architecture\\ type\end{tabular}}} & Hybrid                        & 0.951                        & 0.815                        & 0.892                        & 0.798                        & 0.889                        & 0.879                        & 0.871                        \\ \hline
\multicolumn{1}{|l|}{}                                                                                                      & Fusion                        & 0.760                        & 0.760                        & 0.920                        & 0.660                        & 0.920                        & 1.000                        & 0.837                        \\ \cline{2-9} 
\multicolumn{1}{|l|}{}                                                                                                      & Uni-SS                        & 0.821                        & 0.902                        & 0.860                        & 0.619                        & 0.878                        & 0.668                        & 0.791                        \\ \cline{2-9} 
\multicolumn{1}{|l|}{}                                                                                                      & Multi-SS                      & 0.911                        & 0.742                        & 0.760                        & 0.589                        & 0.855                        & 0.703                        & 0.760                        \\ \cline{2-9} 
\multicolumn{1}{|l|}{}                                                                                                      & Supervised                    & 0.937                        & 0.760                        & 0.727                        & 0.682                        & 0.806                        & 0.789                        & 0.783                        \\ \cline{2-9} 
\multicolumn{1}{|l|}{}                                                                                                      & Multistep                     & 0.867                        & 0.809                        & 0.846                        & 0.893                        & 0.856                        & 0.798                        & 0.845                        \\ \cline{2-9} 
\multicolumn{1}{|l|}{\multirow{-6}{*}{\begin{tabular}[c]{@{}l@{}}Pre-training\\ protocol\end{tabular}}}                     & Robust                        & 0.568                        & 0.641                        & 0.841                        & 0.670                        & 0.685                        & 0.877                        & 0.714                        \\ \hline
\end{tabular}%
}
\vspace{0.5em}
\caption{Summary table of the Mean Absolute Correlation in FFT-50, measured over the $80$ configurations as well as on subsets of configurations based on design choices. }
\label{tab:spearman_corr}
\vspace{-15pt}
\end{table}

\section{Conclusion}

% Among other findings, we find that convolutional architectures perform best for robust fine-tuning in the low-data regime. Despite growing interest in the robustness of attention-based architectures \citep{bhojanapalli2021understanding}, our study suggests they are more difficult to fine-tune for robustness in practice. 
This study reveals a series of actionable insights on the design of robust fine-tuning pipelines in vision classification in the low training data regime. The insights can guide practitioners in selecting more effectively fine-tuning configurations to set-up their own pipeline, speeding up the prototyping process. Aside from guiding practitioners, the insights of this study guide researchers in the field of robust fine-tuning and pre-training. The conclusions identify promising research directions, highlighting strengths and weaknesses of some design choices, and reveal misconceptions regarding the effectiveness of popular design choices in the field.

% Our findings have broader design impacts for vision systems: for example, vision-language models predominantly rely on attention-based backbones \citep{radford2021learning}.

\paragraph{Limitations} 
The insights from this study are contingent to the set of datasets, backbones, and optimization strategies considered. We acknowledge that such insights need to continually evolve with the development of new design choices.  
In this study, the configurations were optimized based on a total compute budget, rather than on an equal number of trials across backbones. This choice reflects the practical reality that some backbones are more challenging to tune due to their compute requirements. This compute-aware tuning approach reflects real-world deployment constraints and promotes energy-conscious model selection \citep{benoit_courty_2024_11171501}. 

% \section*{References}
\bibliographystyle{unsrtnat}
\bibliography{references}

% \newpage
% \section*{NeurIPS Paper Checklist}

% \input{paper_core/checklist}

% \newpage
\appendix

\section{Configurations}
\label{sec:configurations}

Tables \ref{tab:backbone_config_refs} and \ref{tab:configurations} summarize the backbones considered in the study.

\begin{table}
\centering
\scriptsize
\begin{tabular}{|c|c|c|c|c|}
\hline
\textbf{Arch Size} & \textbf{Arch Type}  & \textbf{Pre-training Protocol} & \textbf{Arch Name} & \textbf{HuggingFace ID}                                    \\ \hline
Large              &   Attention     & Self-supervised (unimodal)     & ViT Base           & timm/vit\_base\_patch16\_224.dino                          \\ \hline
Large              &   Attention     & Self-supervised (unimodal)     & ViT Base           & timm/vit\_base\_patch16\_224.mae                           \\ \hline
Large              &   Attention     & Self-supervised (multimodal)   & ViT Base           & timm/vit\_base\_patch16\_clip\_224.laion2b                 \\ \hline
Large              &   Attention     & Supervised                     & ViT Base           & timm/vit\_base\_patch16\_224.augreg\_in21k                 \\ \hline
Large              &   Attention     & Supervised                     & ViT Base           & timm/vit\_base\_patch16\_224.augreg\_in1k                  \\ \hline
Medium             &   Attention     & Supervised                     & ViT Small          & timm/vit\_small\_patch16\_224.augreg\_in21k                \\ \hline
Medium             &   Attention     & Supervised                     & ViT Small          & timm/vit\_small\_patch16\_224.augreg\_in1k                 \\ \hline
Medium             &   Attention     & Supervised                     & DeiT Small         & timm/deit\_small\_patch16\_224.fb\_in1k                    \\ \hline
Large              &   Convolutional & Self-supervised (multimodal)   & ConvNeXt Base      & laion/CLIP-convnext\_base\_w-laion2B-s13B-b82K             \\ \hline
Large              &   Convolutional & Self-supervised (multimodal)   & ConvNeXt Base      & laion/CLIP-convnext\_base\_w-laion\_aesthetic-s13B-b82K    \\ \hline
Large              &   Convolutional & Supervised                     & ConvNeXt Base      & timm/convnext\_base.fb\_in1k                               \\ \hline
Large              &   Convolutional & Supervised                     & ConvNeXt Base      & timm/convnext\_base.fb\_in22k                              \\ \hline
Medium             &   Convolutional & Supervised                     & ConvNeXt Tiny      & timm/convnext\_tiny.fb\_in22k                              \\ \hline
Medium             &   Convolutional & Supervised                     & ConvNeXt Tiny      & timm/convnext\_tiny.fb\_in1k                               \\ \hline
Medium             &   Convolutional & Supervised                     & ResNet50           & timm/resnet50.a1\_in1k                                     \\ \hline
Large              &   Attention     &  Fusion            & ViT Base           & timm/vit\_base\_patch16\_clip\_224.laion2b\_ft\_in1k       \\ \hline
Large              &   Attention     & Supervised (multistep)         & ViT Base           & timm/vit\_base\_patch16\_224.augreg\_in21k\_ft\_in1k       \\ \hline
Medium             &   Attention     & Supervised (multistep)         & ViT Small          & timm/vit\_small\_patch16\_224.augreg\_in21k\_ft\_in1k      \\ \hline
Large              &   Attention     & Self-supervised (unimodal)     & EVA02 Base         & timm/eva02\_base\_patch14\_224.mim\_in22k                  \\ \hline
Medium             &   Attention     & Self-supervised (unimodal)     & EVA02 Tiny         & timm/eva02\_tiny\_patch14\_224.mim\_in22k                  \\ \hline
Large              &   Attention     & Supervised (multistep)         & Swin Base          & timm/swin\_base\_patch4\_window7\_224.ms\_in22k\_ft\_in1k  \\ \hline
Medium             &   Attention     & Supervised                     & Swin Tiny          & timm/swin\_tiny\_patch4\_window7\_224.ms\_in1k             \\ \hline
Large              &   Convolutional & Fusion            & ConvNeXt Base      & timm/convnext\_base.clip\_laion2b\_augreg\_ft\_in12k\_in1k \\ \hline
Large              &   Convolutional & Supervised (multistep)         & ConvNeXt Base      & timm/convnext\_base.fb\_in22k\_ft\_in1k                    \\ \hline
Medium             &   Convolutional & Supervised (multistep)         & ConvNeXt Tiny      & timm/convnext\_tiny.fb\_in22k\_ft\_in1k                    \\ \hline
Small              &   Convolutional & Supervised                     & RegNetX-004        & timm/regnetx\_004.pycls\_in1k                              \\ \hline
Small              &   Convolutional & Supervised                     & EfficientNet-B0    & google/efficientnet-b0                                     \\ \hline
Small              &   Attention     & Supervised                     & DeiT Tiny          & timm/deit\_tiny\_patch16\_224.fb\_in1k                     \\ \hline
Small              & Hybrid              & Supervised                     & MobileViT Small    & apple/mobilevit-small                                      \\ \hline
Small              &   Convolutional & Supervised                     & MobileNetV3        & timm/mobilenetv3\_large\_100.ra\_in1k                      \\ \hline
Small              &   Convolutional & Supervised                     & EdgeNeXt Small     & timm/edgenext\_small.usi\_in1k                             \\ \hline
Small              & Hybrid              & Supervised                     & CoaT Tiny          & timm/coat\_tiny.in1k                                       \\ \hline
Medium             & Hybrid              & Supervised                     & CoAtNet-0          & timm/coatnet\_0\_rw\_224.sw\_in1k                          \\ \hline
Large              & Hybrid              & Supervised (multistep)         & CoAtNet-2          & timm/coatnet\_2\_rw\_224.sw\_in12k\_ft\_in1k               \\ \hline
Large              & Hybrid              & Supervised                     & CoAtNet-2          & timm/coatnet\_2\_rw\_224.sw\_in12k                         \\ \hline
Medium             &   Convolutional & Supervised (robust)            & ResNet50           & robustness/resnet50\_robust                                \\ \hline
Large              &   Convolutional & Supervised (robust)            & ConvNeXt Base      & robustness/convnext\_base\_robust                          \\ \hline
Medium             &   Convolutional & Supervised (robust)            & ConvNeXt Tiny      & robustness/convnext\_tiny\_robust                          \\ \hline
Medium             &   Attention     & Supervised (robust)            & DeiT Small         & robustness/deit\_small\_robust                             \\ \hline
Large              &   Attention     & Supervised (robust)            & ViT Base           & robustness/vit\_base\_robust                               \\ \hline
\end{tabular}%
\vspace{0.8em}
\caption{Summary the 40 configurations considered, described with the architecture size, and type, the pre-training protocol, the architecture name and the official HuggingFace identifier.}
\label{tab:configurations}
\end{table}
% \end{landscape}

% \begin{landscape}
% \pagestyle{empty}%
\begin{table}[ht]
\centering
\scriptsize
\begin{tabular}{|c|c|c|}
\hline
\textbf{Configuration Name}           & \textbf{HuggingFace ID}                                    & \textbf{References}                                                                                 \\ \hline
vit\_b,dino,in1k                      & timm/vit\_base\_patch16\_224.dino                          & \cite{caron2021emerging}       \\ \hline
vit\_b,mae,in1k                       & timm/vit\_base\_patch16\_224.mae                           & \cite{he2022masked}                                                              \\ \hline
vit\_b,clip,laion2b                   & timm/vit\_base\_patch16\_clip\_224.laion2b                 & \cite{ilharco_gabriel_2021_5143773}  \\ \hline
vit\_b,sup,in22k                      & timm/vit\_base\_patch16\_224.augreg\_in21k                 & \cite{steiner2022how}                                                            \\ \hline
vit\_b,sup,in1k                       & timm/vit\_base\_patch16\_224.augreg\_in1k                  & \cite{steiner2022how}                                                            \\ \hline
vit\_s,sup,in22k                      & timm/vit\_small\_patch16\_224.augreg\_in21k                & \cite{steiner2022how}                                                            \\ \hline
vit\_s,sup,in1k                       & timm/vit\_small\_patch16\_224.augreg\_in1k                 & \cite{steiner2022how}                                                            \\ \hline
deit\_s,sup,in1k                      & timm/deit\_small\_patch16\_224.fb\_in1k                    & \cite{touvron2021training}                                                                          \\ \hline
convnext\_b,clip,laion2b              & laion/CLIP-convnext\_base\_w-laion2B-s13B-b82K             & \cite{schuhmann2022laionb}       \\ \hline
convnext\_b,clip,laiona               & laion/CLIP-convnext\_base\_w-laion\_aesthetic-s13B-b82K    & \cite{schuhmann2022laionb}       \\ \hline
convnext\_b,sup,in1k                  & timm/convnext\_base.fb\_in1k                               & \cite{liu2022convnet}                                                                               \\ \hline
convnext\_b,sup,in22k                 & timm/convnext\_base.fb\_in22k                              & \cite{liu2022convnet}                                                                               \\ \hline
convnext\_t,sup,in22k                 & timm/convnext\_tiny.fb\_in22k                              & \cite{liu2022convnet}                                                                               \\ \hline
convnext\_t,sup,in1k                  & timm/convnext\_tiny.fb\_in1k                               & \cite{liu2022convnet}                                                                               \\ \hline
resnet50,sup,in1k                     & timm/resnet50.a1\_in1k                                     & \cite{wightman2021resnet}                                                               \\ \hline
vit\_b,clip,laion2b                   & timm/vit\_base\_patch16\_clip\_224.laion2b\_ft\_in1k       & \cite{schuhmann2022laionb} \\ \hline
vit\_b,sup,in22k-in1k                 & timm/vit\_base\_patch16\_224.augreg\_in21k\_ft\_in1k       & \cite{steiner2022how}                                                            \\ \hline
vit\_s,sup,in22k-in1k                 & timm/vit\_small\_patch16\_224.augreg\_in21k\_ft\_in1k      & \cite{steiner2022how}                                                            \\ \hline
eva02\_b,mim,ink22k                   & timm/eva02\_base\_patch14\_224.mim\_in22k                  & \cite{fang2024eva}                                                                      \\ \hline
eva02\_b,mim,ink22k                   & timm/eva02\_tiny\_patch14\_224.mim\_in22k                  & \cite{fang2024eva}                                                                      \\ \hline
swin\_b,sup,ink22k-in1k               & timm/swin\_base\_patch4\_window7\_224.ms\_in22k\_ft\_in1k  & \cite{liu2021swin}                                                                                  \\ \hline
swin\_t,sup,in1k                      & timm/swin\_tiny\_patch4\_window7\_224.ms\_in1k             & \cite{liu2021swin}                                                                                  \\ \hline
convnext\_b,hybrid,laion2b-in12k-in1k & timm/convnext\_base.clip\_laion2b\_augreg\_ft\_in12k\_in1k & \cite{schuhmann2022laionb}                                     \\ \hline
convnext\_b,sup,in22k-in1k            & timm/convnext\_base.fb\_in22k\_ft\_in1k                    & \cite{liu2022convnet}                                                                               \\ \hline
convnext\_t,sup,in22k-in1k            & timm/convnext\_tiny.fb\_in22k\_ft\_in1k                    & \cite{liu2022convnet}                                                                               \\ \hline
regnetx\_004,sup,in1k                 & timm/regnetx\_004.pycls\_in1k                              & \cite{radosavovic2020designing}                                                                     \\ \hline
efficientnet\_b0,sup,in1k             & google/efficientnet-b0                                     & \cite{tan2019efficientnet}                                                                          \\ \hline
deit\_t,sup,in1k                      & timm/deit\_tiny\_patch16\_224.fb\_in1k                     & \cite{touvron2021training}                                                                          \\ \hline
mobilevit\_s,sup,in1k                 & apple/mobilevit-small                                      & \cite{mehta2022mobilevit}                                                                           \\ \hline
mobilenet\_v3,sup,in1k                & timm/mobilenetv3\_large\_100.ra\_in1k                      & \cite{howard2019searching}                                                      \\ \hline
edgenet\_s,sup,in1k                   & timm/edgenext\_small.usi\_in1k                             & \cite{Maaz2022EdgeNeXt}                                                           \\ \hline
coat\_t,sup,in1k                      & timm/coat\_tiny.in1k                                       & \cite{Xu_2021_ICCV}                                                                                 \\ \hline
coatnet\_0,sup,in1k                   & timm/coatnet\_0\_rw\_224.sw\_in1k                          & \cite{dai2021coatnet}                                                                               \\ \hline
coatnet\_2,sup,in12k-in1k             & timm/coatnet\_2\_rw\_224.sw\_in12k\_ft\_in1k               & \cite{dai2021coatnet}                                                                               \\ \hline
coatnet\_2,sup,in12k                  & timm/coatnet\_2\_rw\_224.sw\_in12k                         & \cite{dai2021coatnet}                                                                               \\ \hline
resnet50,rob-sup,in1k                 & robustness/resnet50\_robust                                & \cite{robustness}                                                                       \\ \hline
convnext\_b,rob-sup,in1k              & N/A                                                        & \cite{singh2024revisiting}                                                          \\ \hline
convnext\_t,rob-sup,in1k              & N/A                                                        & \cite{singh2024revisiting}                                                          \\ \hline
deit\_s,rob-sup,in1k                  & N/A                                                        & \cite{singh2024revisiting}                                                       \\ \hline
vit\_b,rob-sup,in1k                   & N/A                                                        & \cite{singh2024revisiting}                                                       \\ \hline
\end{tabular}
\vspace{0.8em}
\caption{Backbone configurations associated with the official HuggingFace ID, and bibliographical references.}
\label{tab:backbone_config_refs}
\end{table}

\section{Additional technical details}
\label{sec:additional_technical}

The summary of the datasets can be found in Table \ref{tab:dataset_splits}.

\begin{table}[ht]
\centering
\caption{Dataset Summary}
\begin{tabular}{lrrrr}
\toprule
\textbf{Dataset} & \textbf{Train} & \textbf{Validation} & \textbf{Test} & \textbf{Nb. Classes} \\
\midrule
Flowers-102                  & 1020  & 1020  & 1000  & 102 \\
UC Merced Land Use           & 1428  & 252   & 420   & 21 \\
Stanford Cars                & 6922  & 1222  & 1000  & 196 \\
Oxford-IIIT Pet              & 3128  & 552   & 1000  & 37 \\
FGVC Aircraft (2013b)        & 3334  & 3333  & 1000  & 100 \\
Caltech101                   & 5899  & 1042  & 1000  & 101 \\
\bottomrule
\end{tabular}
\label{tab:dataset_splits}
\end{table}

\paragraph{Hyper-parameters optimization.}  
To optimize hyper-parameters, two equally valid strategies can be adopted: a) allocating an equal number of trials per configuration, or b) allocating an equal total compute budget per configuration. In this benchmark, model sizes range from 5M to 90M parameters —a 20× difference— which introduces important variability in inference time. Consequently, strategy a) results in large compute differences between configurations. On the contrary strategy b) offers a more practical and equitable framework. Strategy b) also better reflects the constraints faced by practitioners, who must balance model complexity with available compute resources.

We use an ASHAS scheduler based on hyperband \cite{li2018hyperband}. The scheduler minimizes the loss (either TRADES or Classic AT) associated to each specific case. The loss is evaluated on subsets of $25\%$ of the original validation set, obtained with stratified sub-sampling. The hyper-parameters for FFT are $\eta_1, \eta_2, \gamma_1, \gamma_2$ and the use (or not) of a cosine scheduler. The hyper-parameters for LP are $\eta_2, \gamma_2$ and the use (or not) of a cosine scheduler. 
The optimization campaign was designed to explore as many configurations as possible within a fixed compute budget of 140 minutes per configuration, using a maximum of 50 epochs for FFT-50 and LP-50, and 5 epochs for FFT-5.

\paragraph{Fine-tuning strategies}
For fine-tuning and linear probing, we train using using mixed precision and an automated scaler. The AdamW \cite{loshchilov2019decoupled} optimizer is used in all the settings, it is a commonly used choice in prior works \citep{xu2023autolora, liu2023twins, wang2023better}. The batch size is set to $1042$ using gradient accumulation. A cosine scheduler decaying the learning rate to $0$ across the total number of epochs is considered as hyper-parameter. 

\paragraph{Loss implementation.}
The $\beta$ scalar of TRADES is set to $1$. Classic AT is not a tunable objective, so introducing to TRADES a tunable $\beta$ would give an unfair advantage in term of optimization freedom. In other fields, optimization objectives based on KL divergence are usually set to $1$ \citep{kingma2013auto} as a baseline.
%Additionally, empirical evidence suggests that performance of TRADES is not very sensitive to variations of $\beta$ \citep{zhang2019theoretically}.

\paragraph{Data augmentations.} 
The same data augmentation is used across all configurations and datasets. 
Augmentations during training include random horizontal flip ($p=0.5)$, color jitter (with 0.25 brightness, contrast, and saturation), and random rotations. All images are resized to $224 \times 224$. These augmentations are standard for image classification \cite{torchvision2016}. Normalization and standardization is added directly to the forward pass of the models, rather than in the data augmentation pipeline, which is a standard practice in robustness \cite{singh2024revisiting, liu2023twins}. 

\paragraph{Compute resources}
All experiments were conducted on NVIDIA V100 GPUs with 16GB memory. The hyperparameter optimization used a single V100 GPU, while training and evaluation were run on 4 V100 GPUs. In total, the project used approximately 7 GPU-years: around 6 GPU-years were spent on prototyping, ablation studies, and early results, and 1 GPU-year was used to produce the final reported results.

\section{Crafting test-time perturbations in practice} 
\label{sec:crafting_practice}

We create the test sets with stratified sub-sampling to 1k observations across $5$ perturbation types. This follows prior works \citet{singh2024revisiting} and is used to limit the computational costs of measuring robust generalization. In both adversarial and common perturbation types, we consider diverse ways of creating the perturbations and an iterator increments for each image in the test set over all the possible perturbations considered. 

\paragraph{Adversarial perturbation types} The test-time perturbations used to evaluate the accuracy are generated with AutoAttack for  across $\tau = \{ 1,2,\infty \}$. AutoAttack \cite{croce2020reliable} creates diverse sets of perturbations based on 4 different types of perturbation strategies based on white-box and black-box assumptions. 

\paragraph{Common perturbation types}
For $\tau = \{ \texttt{common} \}$, we generate perturbations using the package proposed in \cite{michaelis2019dragon}. We consider perturbations such as gaussian noise, shot noise, impulse noise, defocus blur, motion blur, zoom blur, snow, brightness, contrast, elastic transform, pixelate, and jpeg compression. The perturbation intensity is set to 3, with 5 being the maximum intensity.

\section{Additional results}
\label{sec:additional_results}

\begin{figure}[ht]
  \centering
  \begin{subfigure}[t]{0.8\textwidth}
    \centering
    \includegraphics[width=\textwidth]{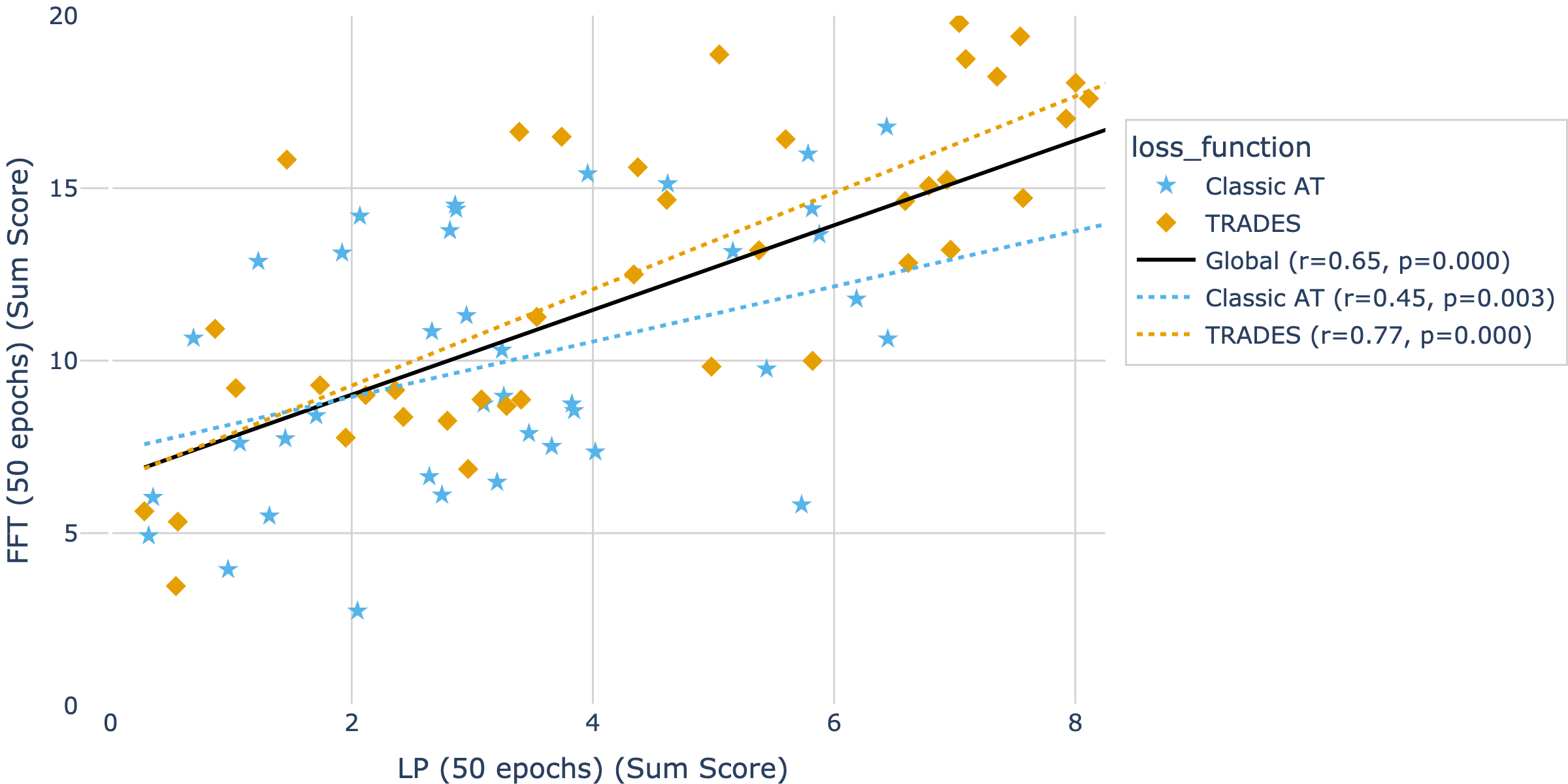}
    \caption{Linear probing (50 epochs).}
    \label{fig:FFT50_LP50}
  \end{subfigure}
  
  \vspace{1em} % add vertical spacing between the figures
  
  \begin{subfigure}[t]{0.8\textwidth}
    \centering
    \includegraphics[width=\textwidth]{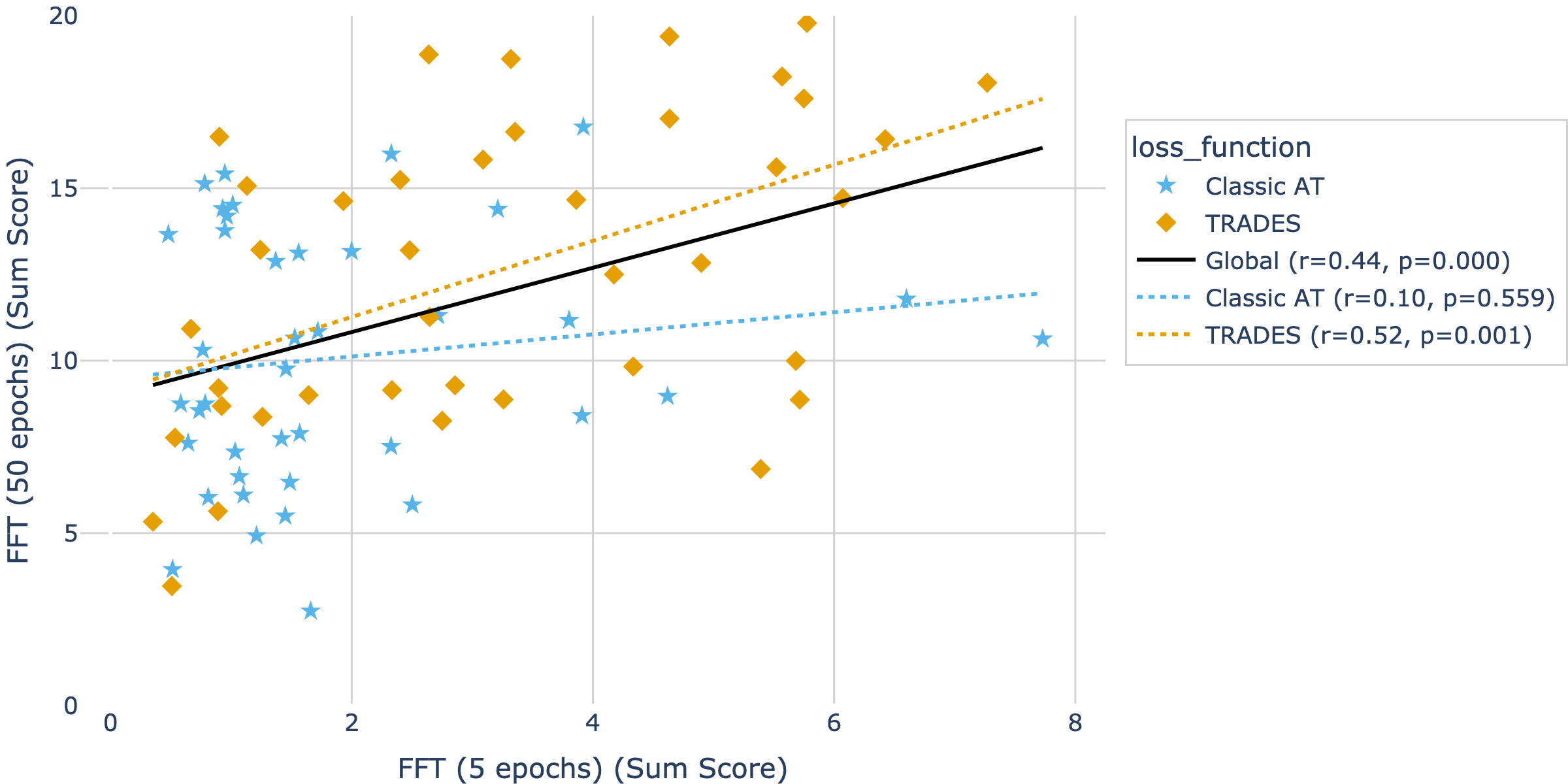}
    \caption{Full fine-tuning (5 epochs).}
    \label{fig:FFT50_FFT5}
  \end{subfigure}

  \caption{Correlation between FFT (50 epochs) and resource constraint fine-tuning settings. The metric is the sum score. The Spearman correlation indicates if the performance in one fine-tuning protocol can predict the performance in the other. The p-values assess the statistical significance of the relationship.}
  \label{fig:comparison}
\end{figure}

Tables \ref{tab:global_ranking_FFT50epochs}, \ref{tab:global_ranking_FFT5epochs} and \ref{tab:global_ranking_LP50epochs} report the global rankings across the fine-tuning protocols considered. Figure \ref{fig:comparison} reports the performance correlation across the fine-tuning protocols considered.

\subsection{FFT-50}

% \paragraph{Modern attention-based architectures outperform vanilla ViTs.} Vanilla ViTs (mean score: $916.15$) are outperformed by recent architectures such as Swin (mean score: $1479.25$) and Eva02 (mean score: $1212.25$), suggesting that the robust fine-tuning community can move past the ViT architecture. In \citet{hua2024initialization}, a comparison between Swin and ViTs was conducted and reported comparable performance
% \aud{Any link to make with results observed in literature?} \maxi{i dont have yet, but open to suggestions.}

\begin{figure}[H]
  \centering
  %--- model size ----------------------------------------------------
  \begin{subfigure}[t]{.24\linewidth}
    \centering
    \includegraphics[width=\linewidth]{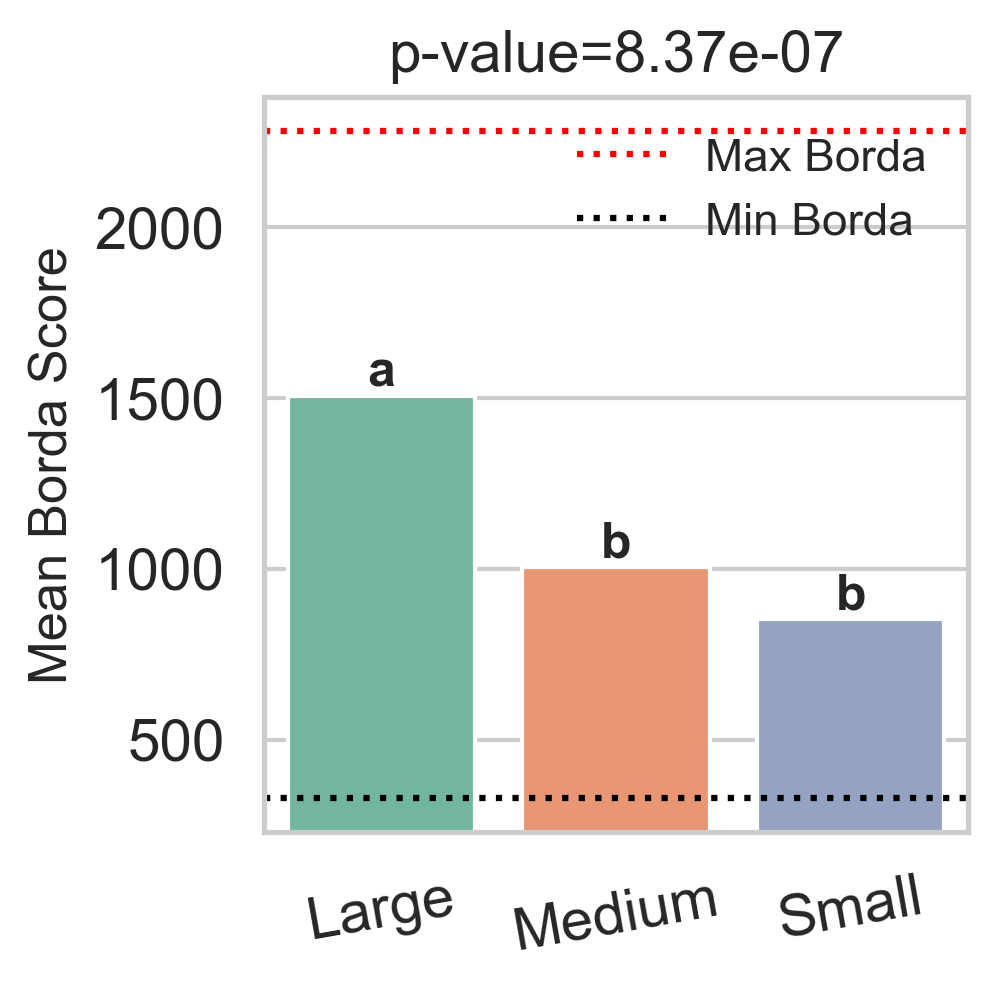}
    \caption{Arch. Scale}
    \label{fig:size}
  \end{subfigure}
  %--- model type ----------------------------------------------------
  \begin{subfigure}[t]{.24\linewidth}
    \centering
    \includegraphics[width=\linewidth]{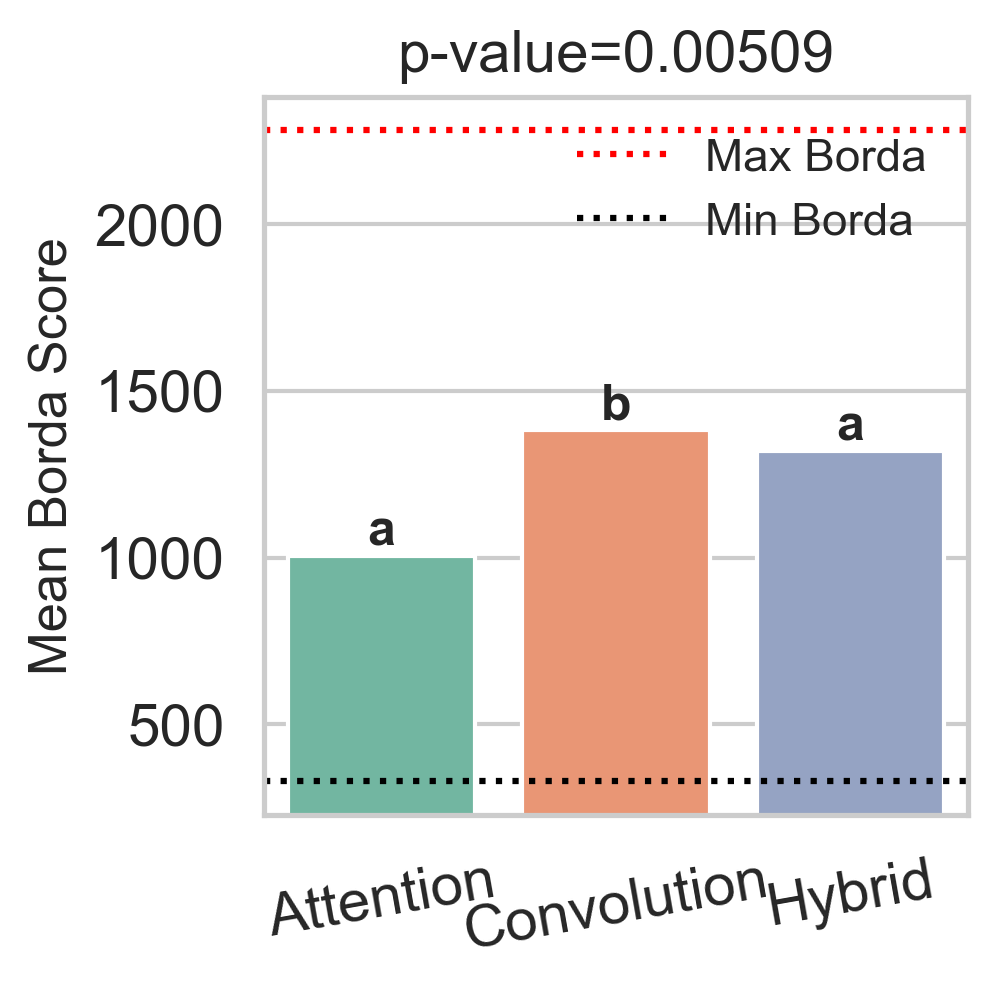}
    \caption{Model type}
    \label{fig:type}
  \end{subfigure}
  %--- loss function -------------------------------------------------
  \begin{subfigure}[t]{.24\linewidth}
    \centering
    \includegraphics[width=\linewidth]{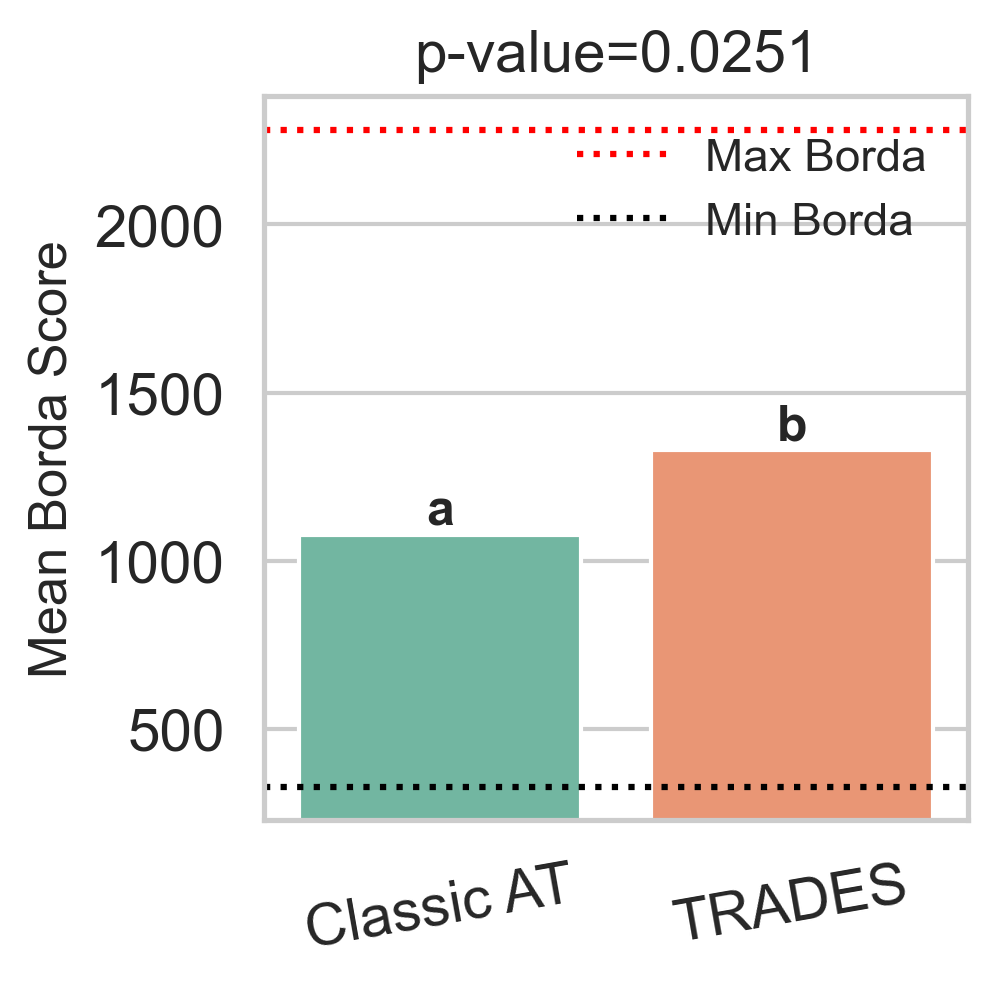}
    \caption{Loss function}
    \label{fig:loss}
  \end{subfigure}
  %--- pre-training strategy ----------------------------------------
  \begin{subfigure}[t]{.24\linewidth}
    \centering
    \includegraphics[width=\linewidth]{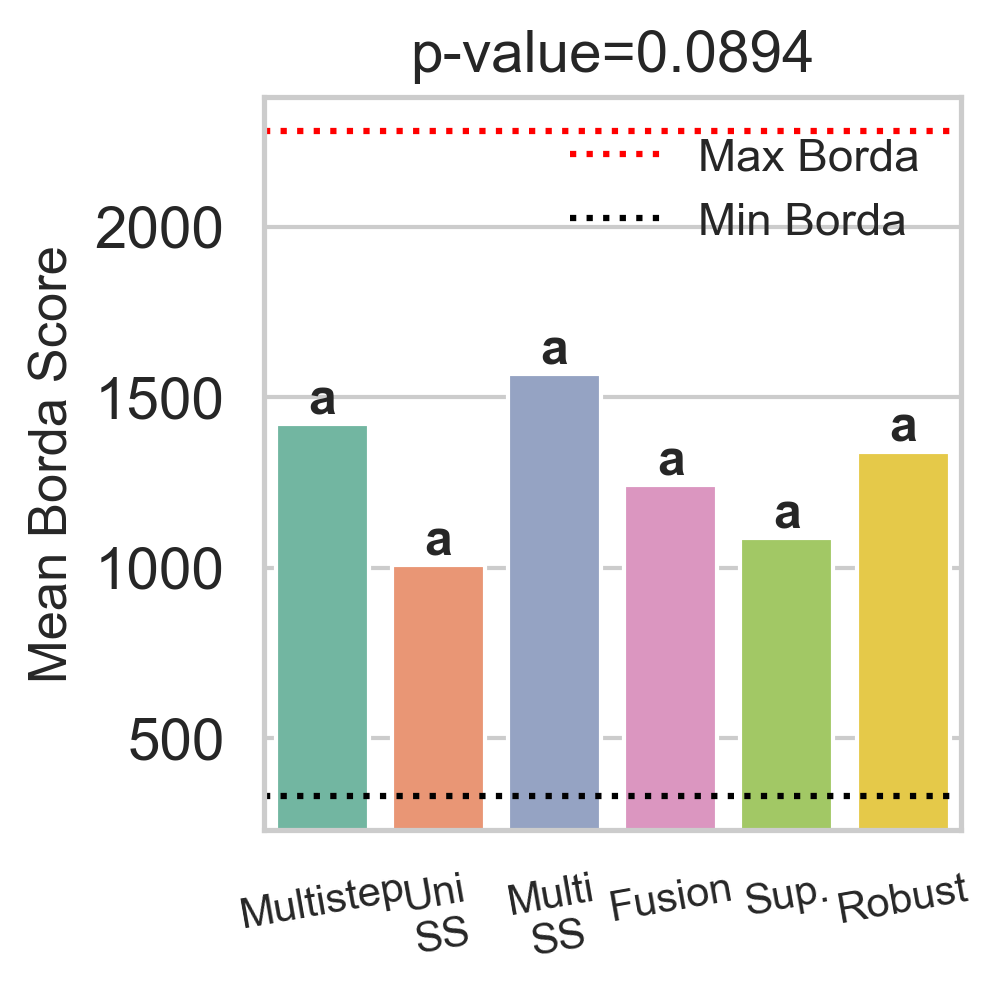}
    \caption{Pre-training strategy}
    \label{fig:pretrain}
  \end{subfigure}

  \caption{Main-effect (Welch’s ANOVA) of mean Borda scores (p-value on top).  
  Post-hoc groupings from Tukey HSD tests are annotated with letters above the bars: bars with different letters belong to significantly different groups, at the $90\%$ confidence level. Results for FFT-50.}
  \label{fig:main_effects}
\end{figure}

\begin{figure}[H]
    \centering
    \includegraphics[width=0.4 \linewidth]{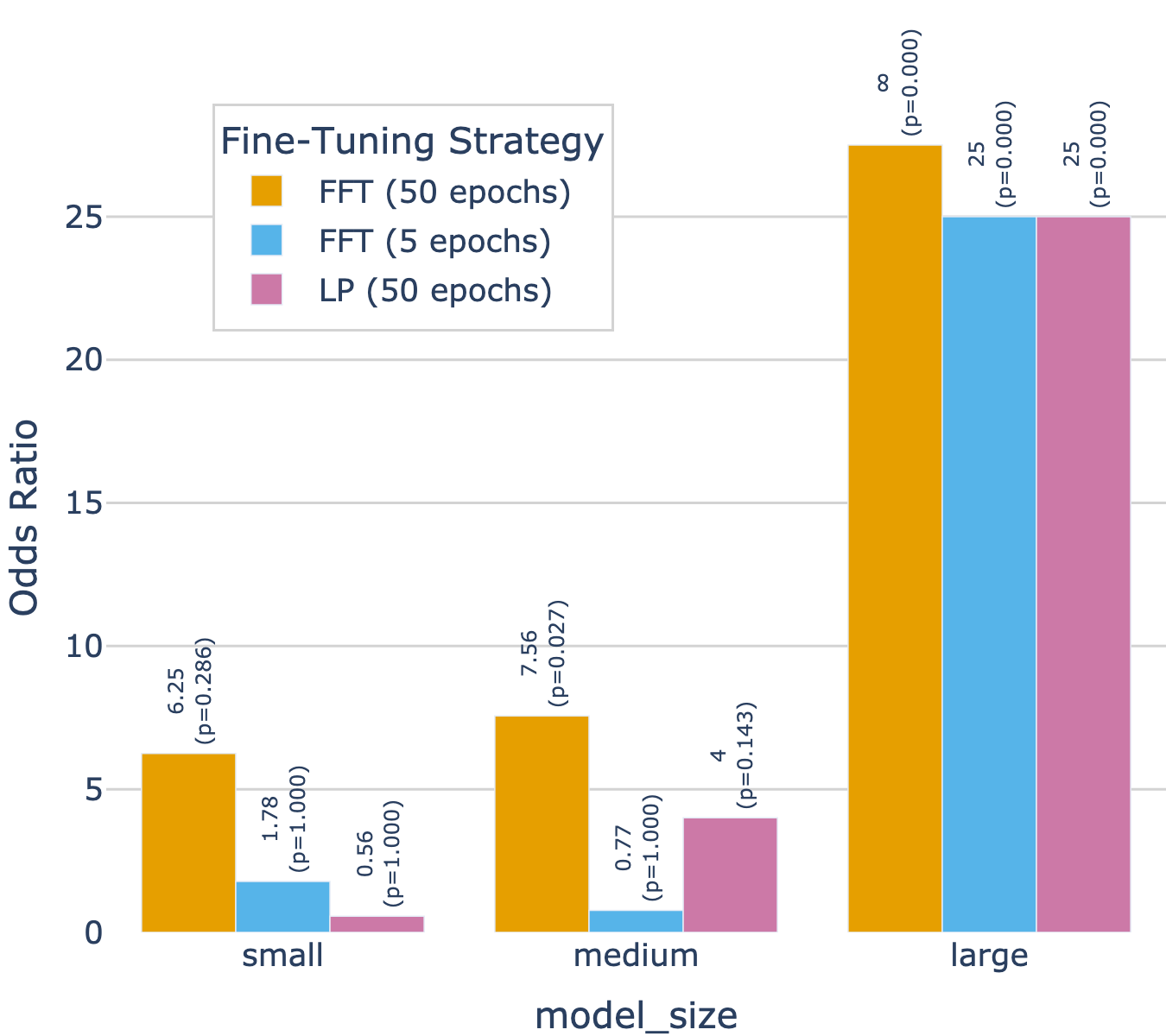}
    \caption{Odds ratio of TRADES outperforming Classic AT, as a function of architecture scale (small, medium, large) and fine-tuning strategy. The odds ratio quantifies how likely TRADES is to improve performance over Classic AT. An odds ratio of 1 indicates no association; values above 1 suggest TRADES tends to outperform Classic AT, and values below 1 suggest the opposite. P-values from Fisher’s Exact Test are shown above each bar. This test evaluates the null hypothesis that TRADES and Classic AT outcomes are independent. }
    \label{fig:TRADES_odds}
\end{figure}

\begin{figure}[ht]
  \centering
  \includegraphics[width=0.5\linewidth]{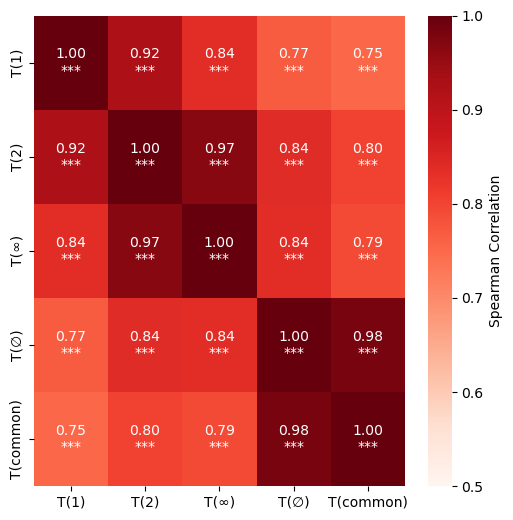}
  \caption{Correlation analysis across robustness metrics in FFT-50, all datasets considered. The heatmap (right) shows Spearman rank correlations with significance levels indicated by stars: * for $p<0.1$, ** for $p<0.05$, and *** for $p<0.01$. Performance on unperturbed observations (i.e., T($\emptyset$) ) is most correlated with robustness to common perturbations (i.e, T(common) ) and least correlated with adversarial perturbations bounded by the $\ell_1$-norm (i.e., T(1) ).}
  \label{fig:correlation_summary}
\end{figure}

% \paragraph{Generalizing to $\ell_1$ perturbations is difficult, and all datasets are not equal.}
% We measure the correlation across the $5$ perturbation types considered (see Figure \ref{fig:correlation_summary}). 
% Generalization to unperturbed observations is most correlated with generalization to common perturbations, and is least correlated to generalization to $\ell_1$ perturbations.  The difficulty of generalizing to $L_1$ perturbations is particularly strong on Oxford-iit, Stanford Cars and Aircraft datasets where even the best configurations exhibits below $0.2$ performance on $\ell_1$, while $\ell_1$ reaches above $0.8$ performance on Caltech101 and LandUse datasets (see Figure \ref{fig:best_per_dataset}). 

\subsection{Full fine tuning 5 epochs}

\begin{table}[ht]
\centering
\resizebox{\textwidth}{!}{%
\begin{tabular}{|c|c|c|c|}
\toprule
\textbf{Size} &                                                                          \textbf{Gold (1st)} &                                                                          \textbf{Silver (2nd)} &                                                                            \textbf{Bronze (3rd)} \\
\midrule
        small &    \makecell{edgenetx\_s,sup,in1k, TRADES \\[0.3ex] \footnotesize (GR:16, BS:1319, SS:3.86)} & \makecell{mobilevit\_s,sup,in1k, Classic AT \\[0.3ex] \footnotesize (GR:18, BS:1294, SS:1.21)} &    \makecell{edgenetx\_s,sup,in1k, Classic AT \\[0.3ex] \footnotesize (GR:34, BS:1113, SS:1.37)} \\
       medium & \makecell{convnext\_t,rob-sup,in1k, TRADES \\[0.3ex] \footnotesize (GR:1, BS:1987, SS:6.07)} &   \makecell{resnet50,rob-sup,in1k, Classic AT \\[0.3ex] \footnotesize (GR:2, BS:1770, SS:6.6)} & \makecell{convnext\_t,rob-sup,in1k, Classic AT \\[0.3ex] \footnotesize (GR:3, BS:1761, SS:7.73)} \\
        large &      \makecell{vit\_b,sup,in1k, Classic AT \\[0.3ex] \footnotesize (GR:6, BS:1502, SS:2.72)} &   \makecell{convnext\_b,clip,laion2b, TRADES \\[0.3ex] \footnotesize (GR:7, BS:1499, SS:4.63)} &     \makecell{convnext\_b,rob-sup,in1k, TRADES \\[0.3ex] \footnotesize (GR:8, BS:1474, SS:3.35)} \\
\bottomrule
\end{tabular}
}
\vspace{0.5em}
\caption{Top fine‑tuning configurations in FFT (5 epochs), with global ranking (GR) based on Borda score (BS), Sum score (SS) also reported below. }
\label{tab:FFT_5_epochs}
\end{table}

\subsection{Linear probing 50 epochs}

\begin{table}[ht]
\centering
\resizebox{\textwidth}{!}{%
\begin{tabular}{|c|c|c|c|}
\toprule
\textbf{Size} &                                                                               \textbf{Gold (1st)} &                                                                            \textbf{Silver (2nd)} &                                                                          \textbf{Bronze (3rd)} \\
\midrule
        small & \makecell{efficientnet\_b0,sup,in1k, Classic AT \\[0.3ex] \footnotesize (GR:9, BS:1435, SS:0.98)} &   \makecell{efficientnet\_b0,sup,in1k, TRADES \\[0.3ex] \footnotesize (GR:10, BS:1387, SS:0.54)} & \makecell{mobilenet\_v3,sup,in1k, Classic AT \\[0.3ex] \footnotesize (GR:43, BS:613, SS:3.83)} \\
       medium &      \makecell{deit\_s,rob-sup,in1k, Classic AT \\[0.3ex] \footnotesize (GR:1, BS:1889, SS:9.13)} &         \makecell{deit\_s,rob-sup,in1k, TRADES \\[0.3ex] \footnotesize (GR:2, BS:1725, SS:6.97)} &   \makecell{convnext\_t,rob-sup,in1k, TRADES \\[0.3ex] \footnotesize (GR:3, BS:1667, SS:7.57)} \\
        large &      \makecell{convnext\_b,rob-sup,in1k, TRADES \\[0.3ex] \footnotesize (GR:6, BS:1484, SS:3.39)} & \makecell{convnext\_b,rob-sup,in1k, Classic AT \\[0.3ex] \footnotesize (GR:8, BS:1459, SS:2.81)} &     \makecell{swin\_b,sup,ink22k-in1k, TRADES \\[0.3ex] \footnotesize (GR:11, BS:967, SS:8.0)} \\
\bottomrule
\end{tabular}
}
\vspace{0.5em}
\caption{Top fine‑tuning configurations in LP (50 epochs), with global ranking (GR) based on Borda score (BS), Sum score (SS) also reported below. }
\label{tab:LP_50_epochs}
\end{table}

% \begin{figure}[ht]
%   \centering
%   \begin{subfigure}[t]{0.49\textwidth}  % ~800 px wide at ~4.2 in
%     \centering
%     \includegraphics[width=\textwidth]{paper_figures/low_cost_proxy_LP_(50_epochs).png}
%     \caption{Linear probing (50 epochs).}
%     \label{fig:FFT50_LP50}
%   \end{subfigure}
%   \hfill
%   \begin{subfigure}[t]{0.49\textwidth}  % ~488 px wide at ~2.54 in
%     \centering
%     \includegraphics[width=\textwidth]{paper_figures/low_cost_proxy_FFT_(5_epochs).png}
%     \caption{Full fine-tuning (5 epochs).}
%     \label{fig:FFT50_FFT5}
%   \end{subfigure}
%   \caption{Correlation between FFT (50 epochs) and resource constraint fine-tuning settings. The metric is the sum score. The Spearman correlation if the performance in one fine-tuning protocol can predict the perform in the other one. The p-values asses the statistical significance of the relationship. }
%   \label{fig:comparison}
% \end{figure}

\begin{table}[htbp]
\centering
\caption{Ranking table for FFT-50}
\label{tab:global_ranking_FFT50epochs}
\resizebox{\textwidth}{!}{%
    \begin{tabular}{lllllrrrr}
\toprule
                       backbone\_name & loss\_function &        pre\_training\_strategy &          model\_type & model\_size &  borda &  score\_sum &  rank\_borda &  rank\_sum \\
\midrule
                convnext\_b,sup,in22k &        TRADES &                   supervised & fully convolutional &      large & 2281.0 &  19.789302 &         1.0 &       1.0 \\
            coatnet\_2,sup,in12k-in1k &        TRADES &       supervised (multistep) &              hybrid &      large & 2127.0 &  18.743905 &         2.0 &       4.0 \\
                 coatnet\_2,sup,in12k &        TRADES &                   supervised &              hybrid &      large & 2116.0 &  18.874429 &         3.0 &       3.0 \\
             convnext\_b,clip,laion2b &        TRADES & self-supervised (multimodal) & fully convolutional &      large & 2067.0 &  19.396968 &         4.0 &       2.0 \\
           convnext\_b,sup,in22k-in1k &        TRADES &       supervised (multistep) & fully convolutional &      large & 2001.0 &  17.016397 &         5.0 &       8.0 \\
                  eva02\_b,mim,ink22k &        TRADES &              self-supervised &     fully attention &      large & 1995.0 &  17.599524 &         6.0 &       7.0 \\
              convnext\_b,clip,laiona &        TRADES & self-supervised (multimodal) & fully convolutional &      large & 1992.0 &  18.235889 &         7.0 &       5.0 \\
              swin\_b,sup,ink22k-in1k &        TRADES &       supervised (multistep) &     fully attention &      large & 1990.0 &  18.053905 &         8.0 &       6.0 \\
             convnext\_b,rob-sup,in1k &        TRADES &          supervised (robust) & fully convolutional &      large & 1974.0 &  16.631302 &         9.0 &      10.0 \\
                 convnext\_b,sup,in1k &        TRADES &                   supervised & fully convolutional &      large & 1938.0 &  16.420429 &        10.0 &      12.0 \\
                convnext\_b,sup,in22k &    Classic AT &                   supervised & fully convolutional &      large & 1922.0 &  15.992889 &        11.0 &      13.0 \\
           convnext\_b,sup,in22k-in1k &    Classic AT &       supervised (multistep) & fully convolutional &      large & 1799.0 &  15.127683 &        12.0 &      18.0 \\
              swin\_b,sup,ink22k-in1k &    Classic AT &       supervised (multistep) &     fully attention &      large & 1792.0 &  16.772365 &        13.0 &       9.0 \\
           convnext\_t,sup,in22k-in1k &        TRADES &       supervised (multistep) & fully convolutional &     medium & 1773.0 &  16.489968 &        14.0 &      11.0 \\
                 convnext\_t,sup,in1k &        TRADES &                   supervised & fully convolutional &     medium & 1681.0 &  15.604810 &        15.0 &      15.0 \\
convnext\_b,hybrid,laion2b-in12k-in1k &        TRADES &                       fusion & fully convolutional &      large & 1656.0 &  14.625873 &        16.0 &      22.0 \\
                 convnext\_b,sup,in1k &    Classic AT &                   supervised & fully convolutional &      large & 1655.0 &  14.390857 &        17.0 &      25.0 \\
             convnext\_b,rob-sup,in1k &    Classic AT &          supervised (robust) & fully convolutional &      large & 1653.0 &  13.771698 &        18.0 &      27.0 \\
                     coat\_t,sup,in1k &        TRADES &                   supervised &              hybrid &      small & 1653.0 &  15.828016 &        18.0 &      14.0 \\
                convnext\_t,sup,in22k &        TRADES &                   supervised & fully convolutional &     medium & 1650.0 &  15.066190 &        20.0 &      19.0 \\
             convnext\_b,clip,laion2b &    Classic AT & self-supervised (multimodal) & fully convolutional &      large & 1634.0 &  15.419238 &        21.0 &      16.0 \\
                  vit\_b,clip,laion2b &        TRADES &                       fusion &     fully attention &      large & 1572.0 &  15.240730 &        22.0 &      17.0 \\
                 edgenetx\_s,sup,in1k &        TRADES &                   supervised &              hybrid &      small & 1552.0 &  14.662857 &        23.0 &      21.0 \\
            coatnet\_2,sup,in12k-in1k &    Classic AT &       supervised (multistep) &              hybrid &      large & 1544.0 &  14.404333 &        24.0 &      24.0 \\
                 deit\_s,rob-sup,in1k &        TRADES &          supervised (robust) &     fully attention &     medium & 1539.0 &  13.210857 &        25.0 &      29.0 \\
             convnext\_t,rob-sup,in1k &        TRADES &          supervised (robust) & fully convolutional &     medium & 1495.0 &  14.712079 &        26.0 &      20.0 \\
              convnext\_b,clip,laiona &    Classic AT & self-supervised (multimodal) & fully convolutional &      large & 1478.0 &  14.510619 &        27.0 &      23.0 \\
                 convnext\_t,sup,in1k &    Classic AT &                   supervised & fully convolutional &     medium & 1471.0 &  14.190667 &        28.0 &      26.0 \\
                 coatnet\_2,sup,in12k &    Classic AT &                   supervised &              hybrid &      large & 1464.0 &  13.656476 &        29.0 &      28.0 \\
                  vit\_b,clip,laion2b &        TRADES & self-supervised (multimodal) &     fully attention &      large & 1439.0 &  12.831397 &        30.0 &      34.0 \\
                  eva02\_b,mim,ink22k &    Classic AT &              self-supervised &     fully attention &      large & 1425.0 &  13.163381 &        31.0 &      31.0 \\
                convnext\_t,sup,in22k &    Classic AT &                   supervised & fully convolutional &     medium & 1394.0 &  13.121667 &        32.0 &      32.0 \\
                 edgenetx\_s,sup,in1k &    Classic AT &                   supervised &              hybrid &      small & 1356.0 &  12.877714 &        33.0 &      33.0 \\
                  coatnet\_0,sup,in1k &        TRADES &                   supervised &              hybrid &     medium & 1287.0 &  13.198714 &        34.0 &      30.0 \\
                      vit\_b,sup,in1k &        TRADES &                   supervised &     fully attention &      large & 1267.0 &  12.497778 &        35.0 &      35.0 \\
               resnet50,rob-sup,in1k &    Classic AT &          supervised (robust) & fully convolutional &     medium & 1255.0 &  11.786952 &        36.0 &      36.0 \\
                 deit\_s,rob-sup,in1k &    Classic AT &          supervised (robust) &     fully attention &     medium & 1194.0 &  11.169270 &        37.0 &      39.0 \\
                      vit\_b,mae,in1k &        TRADES &              self-supervised &     fully attention &      large & 1176.0 &  10.922762 &        38.0 &      40.0 \\
                     swin\_t,sup,in1k &        TRADES &                   supervised &     fully attention &     medium & 1153.0 &   9.993524 &        39.0 &      45.0 \\
             convnext\_t,rob-sup,in1k &    Classic AT &          supervised (robust) & fully convolutional &     medium & 1124.0 &  10.630048 &        40.0 &      43.0 \\
                  vit\_b,rob-sup,in1k &        TRADES &          supervised (robust) &     fully attention &      large & 1122.0 &   9.284619 &        41.0 &      48.0 \\
convnext\_b,hybrid,laion2b-in12k-in1k &    Classic AT &                       fusion & fully convolutional &      large & 1114.0 &   9.757952 &        42.0 &      47.0 \\
                      vit\_b,sup,in1k &    Classic AT &                   supervised &     fully attention &      large & 1086.0 &  11.306857 &        43.0 &      37.0 \\
                      vit\_b,mae,in1k &    Classic AT &              self-supervised &     fully attention &      large & 1049.0 &  10.652857 &        44.0 &      42.0 \\
               resnet50,rob-sup,in1k &        TRADES &          supervised (robust) & fully convolutional &     medium & 1035.0 &   9.827397 &        45.0 &      46.0 \\
                  vit\_b,rob-sup,in1k &    Classic AT &          supervised (robust) &     fully attention &      large & 1004.0 &   7.738222 &        46.0 &      64.0 \\
                     swin\_t,sup,in1k &    Classic AT &                   supervised &     fully attention &     medium &  984.0 &  10.305952 &        47.0 &      44.0 \\
                     deit\_t,sup,in1k &        TRADES &                   supervised &     fully attention &      small &  973.0 &  11.266238 &        48.0 &      38.0 \\
                regnetx\_004,sup,in1k &    Classic AT &                   supervised & fully convolutional &      small &  930.0 &  10.845762 &        49.0 &      41.0 \\
           convnext\_t,sup,in22k-in1k &    Classic AT &       supervised (multistep) & fully convolutional &     medium &  922.0 &   8.969159 &        50.0 &      52.0 \\
                     deit\_s,sup,in1k &    Classic AT &                   supervised &     fully attention &     medium &  917.0 &   8.404937 &        51.0 &      59.0 \\
                     vit\_b,sup,in22k &        TRADES &                   supervised &     fully attention &      large &  909.0 &   9.204524 &        52.0 &      49.0 \\
                     vit\_b,sup,in22k &    Classic AT &                   supervised &     fully attention &      large &  881.0 &   8.744079 &        53.0 &      56.0 \\
                  coatnet\_0,sup,in1k &    Classic AT &                   supervised &              hybrid &     medium &  846.0 &   7.354095 &        54.0 &      67.0 \\
               vit\_b,sup,ink22k-in1k &    Classic AT &       supervised (multistep) &     fully attention &      large &  840.0 &   7.513095 &        55.0 &      66.0 \\
               mobilenet\_v3,sup,in1k &    Classic AT &                   supervised & fully convolutional &      small &  832.0 &   8.747540 &        56.0 &      55.0 \\
               vit\_b,sup,ink22k-in1k &        TRADES &       supervised (multistep) &     fully attention &      large &  825.0 &   7.766048 &        57.0 &      63.0 \\
                     vit\_s,sup,in22k &        TRADES &                   supervised &     fully attention &     medium &  823.0 &   8.256286 &        58.0 &      61.0 \\
                  eva02\_b,mim,ink22k &        TRADES &              self-supervised &     fully attention &     medium &  804.0 &   8.368905 &        59.0 &      60.0 \\
                  vit\_b,clip,laion2b &    Classic AT & self-supervised (multimodal) &     fully attention &      large &  791.0 &   8.554016 &        60.0 &      58.0 \\
               vit\_s,sup,ink22k-in1k &        TRADES &       supervised (multistep) &     fully attention &     medium &  775.0 &   8.685698 &        61.0 &      57.0 \\
                     coat\_t,sup,in1k &    Classic AT &                   supervised &              hybrid &      small &  734.0 &   6.036095 &        62.0 &      72.0 \\
               mobilenet\_v3,sup,in1k &        TRADES &                   supervised & fully convolutional &      small &  731.0 &   8.869429 &        63.0 &      54.0 \\
                   resnet50,sup,in1k &    Classic AT &                   supervised & fully convolutional &     medium &  726.0 &   7.891190 &        64.0 &      62.0 \\
                   resnet50,sup,in1k &        TRADES &                   supervised & fully convolutional &     medium &  692.0 &   8.874571 &        65.0 &      53.0 \\
                      vit\_s,sup,in1k &        TRADES &                   supervised &     fully attention &     medium &  671.0 &   9.144730 &        66.0 &      50.0 \\
                     vit\_s,sup,in22k &    Classic AT &                   supervised &     fully attention &     medium &  666.0 &   6.476619 &        67.0 &      70.0 \\
               vit\_s,sup,ink22k-in1k &    Classic AT &       supervised (multistep) &     fully attention &     medium &  659.0 &   6.104857 &        68.0 &      71.0 \\
                  eva02\_b,mim,ink22k &    Classic AT &              self-supervised &     fully attention &     medium &  627.0 &   5.497159 &        69.0 &      75.0 \\
                     deit\_t,sup,in1k &    Classic AT &                   supervised &     fully attention &      small &  626.0 &   7.611524 &        70.0 &      65.0 \\
                  vit\_b,clip,laion2b &    Classic AT &                       fusion &     fully attention &      large &  625.0 &   5.820127 &        71.0 &      73.0 \\
                regnetx\_004,sup,in1k &        TRADES &                   supervised & fully convolutional &      small &  624.0 &   9.000444 &        72.0 &      51.0 \\
                     vit\_b,dino,in1k &    Classic AT &              self-supervised &     fully attention &     medium &  611.0 &   6.640190 &        73.0 &      69.0 \\
                mobilevit\_s,sup,in1k &    Classic AT &                   supervised &              hybrid &      small &  610.0 &   4.924286 &        74.0 &      77.0 \\
                     deit\_s,sup,in1k &        TRADES &                   supervised &     fully attention &     medium &  586.0 &   6.858127 &        75.0 &      68.0 \\
                mobilevit\_s,sup,in1k &        TRADES &                   supervised &              hybrid &      small &  554.0 &   5.632667 &        76.0 &      74.0 \\
                      vit\_s,sup,in1k &    Classic AT &                   supervised &     fully attention &     medium &  433.0 &   2.742905 &        77.0 &      80.0 \\
            efficientnet\_b0,sup,in1k &    Classic AT &                   supervised & fully convolutional &      small &  423.0 &   3.946429 &        78.0 &      78.0 \\
                     vit\_b,dino,in1k &        TRADES &              self-supervised &     fully attention &     medium &  367.0 &   5.332063 &        79.0 &      76.0 \\
            efficientnet\_b0,sup,in1k &        TRADES &                   supervised & fully convolutional &      small &  329.0 &   3.465190 &        80.0 &      79.0 \\
\bottomrule
\end{tabular}
}
\end{table}

\begin{table}[htbp]
\centering
\caption{Ranking table for full fine tuning (5 epochs)}
\label{tab:global_ranking_FFT5epochs}
\resizebox{\textwidth}{!}{%
    \begin{tabular}{lllllrrrr}
\toprule
                       backbone\_name & loss\_function &        pre\_training\_strategy &          model\_type & model\_size &  borda &  score\_sum &  rank\_borda &  rank\_sum \\
\midrule
             convnext\_t,rob-sup,in1k &        TRADES &          supervised (robust) & fully convolutional &     medium & 1987.0 &   6.071905 &         1.0 &       5.0 \\
               resnet50,rob-sup,in1k &    Classic AT &          supervised (robust) & fully convolutional &     medium & 1770.0 &   6.600286 &         2.0 &       3.0 \\
             convnext\_t,rob-sup,in1k &    Classic AT &          supervised (robust) & fully convolutional &     medium & 1761.0 &   7.729302 &         3.0 &       1.0 \\
               resnet50,rob-sup,in1k &        TRADES &          supervised (robust) & fully convolutional &     medium & 1746.0 &   4.334524 &         4.0 &      17.0 \\
                     deit\_s,sup,in1k &    Classic AT &                   supervised &     fully attention &     medium & 1508.0 &   3.909571 &         5.0 &      20.0 \\
                      vit\_b,sup,in1k &    Classic AT &                   supervised &     fully attention &      large & 1502.0 &   2.715952 &         6.0 &      30.0 \\
             convnext\_b,clip,laion2b &        TRADES & self-supervised (multimodal) & fully convolutional &      large & 1499.0 &   4.634952 &         7.0 &      15.0 \\
             convnext\_b,rob-sup,in1k &        TRADES &          supervised (robust) & fully convolutional &      large & 1474.0 &   3.354571 &         8.0 &      23.0 \\
                convnext\_t,sup,in22k &    Classic AT &                   supervised & fully convolutional &     medium & 1442.0 &   1.559492 &         9.0 &      45.0 \\
convnext\_b,hybrid,laion2b-in12k-in1k &        TRADES &                       fusion & fully convolutional &      large & 1438.0 &   1.930810 &        10.0 &      40.0 \\
              convnext\_b,clip,laiona &        TRADES & self-supervised (multimodal) & fully convolutional &      large & 1397.0 &   5.568825 &        11.0 &      10.0 \\
                     vit\_s,sup,in22k &    Classic AT &                   supervised &     fully attention &     medium & 1358.0 &   1.487381 &        12.0 &      47.0 \\
              swin\_b,sup,ink22k-in1k &        TRADES &       supervised (multistep) &     fully attention &      large & 1340.0 &   7.269683 &        13.0 &       2.0 \\
                  vit\_b,clip,laion2b &    Classic AT & self-supervised (multimodal) &     fully attention &      large & 1327.0 &   0.736333 &        14.0 &      72.0 \\
                  vit\_b,clip,laion2b &        TRADES & self-supervised (multimodal) &     fully attention &      large & 1321.0 &   4.898714 &        15.0 &      13.0 \\
                 edgenetx\_s,sup,in1k &        TRADES &                   supervised &              hybrid &      small & 1319.0 &   3.861762 &        16.0 &      21.0 \\
              convnext\_b,clip,laiona &    Classic AT & self-supervised (multimodal) & fully convolutional &      large & 1304.0 &   1.013429 &        17.0 &      59.0 \\
                mobilevit\_s,sup,in1k &    Classic AT &                   supervised &              hybrid &      small & 1294.0 &   1.210571 &        18.0 &      54.0 \\
                      vit\_s,sup,in1k &    Classic AT &                   supervised &     fully attention &     medium & 1294.0 &   1.659968 &        18.0 &      42.0 \\
                  eva02\_b,mim,ink22k &    Classic AT &              self-supervised &     fully attention &     medium & 1288.0 &   1.448952 &        20.0 &      49.0 \\
                  eva02\_b,mim,ink22k &        TRADES &              self-supervised &     fully attention &      large & 1279.0 &   5.749286 &        21.0 &       7.0 \\
                 deit\_s,rob-sup,in1k &    Classic AT &          supervised (robust) &     fully attention &     medium & 1270.0 &   3.802190 &        22.0 &      22.0 \\
                  vit\_b,rob-sup,in1k &        TRADES &          supervised (robust) &     fully attention &      large & 1247.0 &   2.857143 &        23.0 &      28.0 \\
             convnext\_b,rob-sup,in1k &    Classic AT &          supervised (robust) & fully convolutional &      large & 1235.0 &   0.948587 &        24.0 &      61.0 \\
              swin\_b,sup,ink22k-in1k &    Classic AT &       supervised (multistep) &     fully attention &      large & 1229.0 &   3.921143 &        25.0 &      19.0 \\
                 deit\_s,rob-sup,in1k &        TRADES &          supervised (robust) &     fully attention &     medium & 1219.0 &   1.242429 &        26.0 &      53.0 \\
                      vit\_b,sup,in1k &        TRADES &                   supervised &     fully attention &      large & 1215.0 &   4.174905 &        27.0 &      18.0 \\
                 convnext\_b,sup,in1k &        TRADES &                   supervised & fully convolutional &      large & 1207.0 &   6.424079 &        28.0 &       4.0 \\
             convnext\_b,clip,laion2b &    Classic AT & self-supervised (multimodal) & fully convolutional &      large & 1179.0 &   0.948095 &        29.0 &      62.0 \\
                     swin\_t,sup,in1k &        TRADES &                   supervised &     fully attention &     medium & 1167.0 &   5.682810 &        30.0 &       9.0 \\
                  eva02\_b,mim,ink22k &    Classic AT &              self-supervised &     fully attention &      large & 1162.0 &   1.999667 &        31.0 &      39.0 \\
convnext\_b,hybrid,laion2b-in12k-in1k &    Classic AT &                       fusion & fully convolutional &      large & 1149.0 &   1.455492 &        32.0 &      48.0 \\
                  vit\_b,clip,laion2b &        TRADES &                       fusion &     fully attention &      large & 1144.0 &   2.402365 &        33.0 &      35.0 \\
                 edgenetx\_s,sup,in1k &    Classic AT &                   supervised &              hybrid &      small & 1113.0 &   1.370095 &        34.0 &      51.0 \\
                  eva02\_b,mim,ink22k &        TRADES &              self-supervised &     fully attention &     medium & 1111.0 &   1.260238 &        35.0 &      52.0 \\
                 convnext\_b,sup,in1k &    Classic AT &                   supervised & fully convolutional &      large & 1110.0 &   3.211460 &        36.0 &      26.0 \\
                  vit\_b,rob-sup,in1k &    Classic AT &          supervised (robust) &     fully attention &      large & 1104.0 &   1.418476 &        37.0 &      50.0 \\
                  vit\_b,clip,laion2b &    Classic AT &                       fusion &     fully attention &      large & 1087.0 &   2.502968 &        38.0 &      33.0 \\
            coatnet\_2,sup,in12k-in1k &    Classic AT &       supervised (multistep) &              hybrid &      large & 1069.0 &   0.930063 &        39.0 &      63.0 \\
                 convnext\_t,sup,in1k &    Classic AT &                   supervised & fully convolutional &     medium & 1049.0 &   0.964524 &        40.0 &      60.0 \\
                     vit\_b,dino,in1k &    Classic AT &              self-supervised &     fully attention &     medium & 1040.0 &   1.068413 &        41.0 &      57.0 \\
               mobilenet\_v3,sup,in1k &        TRADES &                   supervised & fully convolutional &      small & 1040.0 &   5.715381 &        41.0 &       8.0 \\
            coatnet\_2,sup,in12k-in1k &        TRADES &       supervised (multistep) &              hybrid &      large & 1038.0 &   3.320302 &        43.0 &      24.0 \\
                     deit\_t,sup,in1k &        TRADES &                   supervised &     fully attention &      small & 1015.0 &   2.646619 &        44.0 &      31.0 \\
                      vit\_s,sup,in1k &        TRADES &                   supervised &     fully attention &     medium & 1002.0 &   2.334365 &        45.0 &      36.0 \\
           convnext\_b,sup,in22k-in1k &        TRADES &       supervised (multistep) & fully convolutional &      large &  999.0 &   4.635571 &        46.0 &      14.0 \\
           convnext\_t,sup,in22k-in1k &    Classic AT &       supervised (multistep) & fully convolutional &     medium &  995.0 &   4.619159 &        47.0 &      16.0 \\
                 convnext\_t,sup,in1k &        TRADES &                   supervised & fully convolutional &     medium &  985.0 &   5.521524 &        48.0 &      11.0 \\
                     deit\_s,sup,in1k &        TRADES &                   supervised &     fully attention &     medium &  974.0 &   5.391762 &        49.0 &      12.0 \\
                convnext\_b,sup,in22k &    Classic AT &                   supervised & fully convolutional &      large &  971.0 &   2.328603 &        50.0 &      37.0 \\
                     coat\_t,sup,in1k &        TRADES &                   supervised &              hybrid &      small &  964.0 &   3.089048 &        51.0 &      27.0 \\
                     deit\_t,sup,in1k &    Classic AT &                   supervised &     fully attention &      small &  958.0 &   0.643667 &        52.0 &      74.0 \\
               vit\_s,sup,ink22k-in1k &        TRADES &       supervised (multistep) &     fully attention &     medium &  946.0 &   0.921730 &        53.0 &      64.0 \\
               vit\_s,sup,ink22k-in1k &    Classic AT &       supervised (multistep) &     fully attention &     medium &  929.0 &   1.101921 &        54.0 &      56.0 \\
                      vit\_b,mae,in1k &    Classic AT &              self-supervised &     fully attention &      large &  914.0 &   1.527381 &        55.0 &      46.0 \\
                  coatnet\_0,sup,in1k &    Classic AT &                   supervised &              hybrid &     medium &  911.0 &   1.033619 &        56.0 &      58.0 \\
                convnext\_b,sup,in22k &        TRADES &                   supervised & fully convolutional &      large &  906.0 &   5.775333 &        57.0 &       6.0 \\
                convnext\_t,sup,in22k &        TRADES &                   supervised & fully convolutional &     medium &  898.0 &   1.131762 &        58.0 &      55.0 \\
            efficientnet\_b0,sup,in1k &        TRADES &                   supervised & fully convolutional &      small &  877.0 &   0.509143 &        59.0 &      78.0 \\
                regnetx\_004,sup,in1k &    Classic AT &                   supervised & fully convolutional &      small &  864.0 &   1.719254 &        60.0 &      41.0 \\
                  coatnet\_0,sup,in1k &        TRADES &                   supervised &              hybrid &     medium &  799.0 &   2.481000 &        61.0 &      34.0 \\
               vit\_b,sup,ink22k-in1k &    Classic AT &       supervised (multistep) &     fully attention &      large &  797.0 &   2.326381 &        62.0 &      38.0 \\
           convnext\_t,sup,in22k-in1k &        TRADES &       supervised (multistep) & fully convolutional &     medium &  792.0 &   0.902619 &        63.0 &      65.0 \\
            efficientnet\_b0,sup,in1k &    Classic AT &                   supervised & fully convolutional &      small &  728.0 &   0.514667 &        64.0 &      77.0 \\
               vit\_b,sup,ink22k-in1k &        TRADES &       supervised (multistep) &     fully attention &      large &  705.0 &   0.534190 &        65.0 &      76.0 \\
                     swin\_t,sup,in1k &    Classic AT &                   supervised &     fully attention &     medium &  695.0 &   0.764810 &        66.0 &      71.0 \\
                 coatnet\_2,sup,in12k &        TRADES &                   supervised &              hybrid &      large &  694.0 &   2.639000 &        67.0 &      32.0 \\
                     vit\_s,sup,in22k &        TRADES &                   supervised &     fully attention &     medium &  675.0 &   2.750571 &        68.0 &      29.0 \\
                mobilevit\_s,sup,in1k &        TRADES &                   supervised &              hybrid &      small &  662.0 &   0.891159 &        69.0 &      67.0 \\
                     vit\_b,sup,in22k &    Classic AT &                   supervised &     fully attention &      large &  656.0 &   0.785127 &        70.0 &      69.0 \\
                   resnet50,sup,in1k &        TRADES &                   supervised & fully convolutional &     medium &  652.0 &   3.259000 &        71.0 &      25.0 \\
                      vit\_b,mae,in1k &        TRADES &              self-supervised &     fully attention &      large &  608.0 &   0.666905 &        72.0 &      73.0 \\
                     vit\_b,sup,in22k &        TRADES &                   supervised &     fully attention &      large &  605.0 &   0.896143 &        73.0 &      66.0 \\
                     coat\_t,sup,in1k &    Classic AT &                   supervised &              hybrid &      small &  593.0 &   0.810333 &        74.0 &      68.0 \\
                   resnet50,sup,in1k &    Classic AT &                   supervised & fully convolutional &     medium &  581.0 &   1.568095 &        75.0 &      44.0 \\
           convnext\_b,sup,in22k-in1k &    Classic AT &       supervised (multistep) & fully convolutional &      large &  569.0 &   0.780429 &        76.0 &      70.0 \\
               mobilenet\_v3,sup,in1k &    Classic AT &                   supervised & fully convolutional &      small &  568.0 &   0.582571 &        77.0 &      75.0 \\
                regnetx\_004,sup,in1k &        TRADES &                   supervised & fully convolutional &      small &  533.0 &   1.642524 &        78.0 &      43.0 \\
                 coatnet\_2,sup,in12k &    Classic AT &                   supervised &              hybrid &      large &  497.0 &   0.479508 &        79.0 &      79.0 \\
                     vit\_b,dino,in1k &        TRADES &              self-supervised &     fully attention &     medium &  415.0 &   0.351603 &        80.0 &      80.0 \\
\bottomrule
\end{tabular}
}
\end{table}

\begin{table}[htbp]
\centering
\caption{Ranking table for linear probing (50 epochs)}
\label{tab:global_ranking_LP50epochs}
\resizebox{\textwidth}{!}{%
    \begin{tabular}{lllllrrrr}
\toprule
                       backbone\_name & loss\_function &        pre\_training\_strategy &          model\_type & model\_size &  borda &  score\_sum &  rank\_borda &  rank\_sum \\
\midrule
                 deit\_s,rob-sup,in1k &    Classic AT &          supervised (robust) &     fully attention &     medium & 1889.0 &   9.132143 &         1.0 &       1.0 \\
                 deit\_s,rob-sup,in1k &        TRADES &          supervised (robust) &     fully attention &     medium & 1725.0 &   6.966619 &         2.0 &      10.0 \\
             convnext\_t,rob-sup,in1k &        TRADES &          supervised (robust) & fully convolutional &     medium & 1667.0 &   7.566746 &         3.0 &       5.0 \\
             convnext\_t,rob-sup,in1k &    Classic AT &          supervised (robust) & fully convolutional &     medium & 1644.0 &   6.445508 &         4.0 &      15.0 \\
               resnet50,rob-sup,in1k &        TRADES &          supervised (robust) & fully convolutional &     medium & 1529.0 &   4.983190 &         5.0 &      28.0 \\
             convnext\_b,rob-sup,in1k &        TRADES &          supervised (robust) & fully convolutional &      large & 1484.0 &   3.389619 &         6.0 &      42.0 \\
               resnet50,rob-sup,in1k &    Classic AT &          supervised (robust) & fully convolutional &     medium & 1479.0 &   6.186667 &         7.0 &      17.0 \\
             convnext\_b,rob-sup,in1k &    Classic AT &          supervised (robust) & fully convolutional &      large & 1459.0 &   2.814556 &         8.0 &      53.0 \\
            efficientnet\_b0,sup,in1k &    Classic AT &                   supervised & fully convolutional &      small & 1435.0 &   0.975095 &         9.0 &      73.0 \\
            efficientnet\_b0,sup,in1k &        TRADES &                   supervised & fully convolutional &      small & 1387.0 &   0.542524 &        10.0 &      77.0 \\
              swin\_b,sup,ink22k-in1k &        TRADES &       supervised (multistep) &     fully attention &      large &  967.0 &   8.003857 &        11.0 &       3.0 \\
                  eva02\_b,mim,ink22k &    Classic AT &              self-supervised &     fully attention &      large &  953.0 &   5.160714 &        12.0 &      26.0 \\
           convnext\_b,sup,in22k-in1k &        TRADES &       supervised (multistep) & fully convolutional &      large &  949.0 &   7.923714 &        13.0 &       4.0 \\
                  eva02\_b,mim,ink22k &        TRADES &              self-supervised &     fully attention &      large &  948.0 &   8.113413 &        14.0 &       2.0 \\
                convnext\_t,sup,in22k &        TRADES &                   supervised & fully convolutional &     medium &  946.0 &   6.786413 &        15.0 &      12.0 \\
             convnext\_b,clip,laion2b &        TRADES & self-supervised (multimodal) & fully convolutional &      large &  901.0 &   7.543333 &        16.0 &       6.0 \\
            coatnet\_2,sup,in12k-in1k &        TRADES &       supervised (multistep) &              hybrid &      large &  887.0 &   7.091619 &        17.0 &       8.0 \\
                convnext\_b,sup,in22k &        TRADES &                   supervised & fully convolutional &      large &  860.0 &   7.037429 &        18.0 &       9.0 \\
                  vit\_b,clip,laion2b &        TRADES &                       fusion &     fully attention &      large &  835.0 &   6.935571 &        19.0 &      11.0 \\
convnext\_b,hybrid,laion2b-in12k-in1k &        TRADES &                       fusion & fully convolutional &      large &  821.0 &   6.589540 &        20.0 &      14.0 \\
              convnext\_b,clip,laiona &        TRADES & self-supervised (multimodal) & fully convolutional &      large &  818.0 &   7.351238 &        21.0 &       7.0 \\
              swin\_b,sup,ink22k-in1k &    Classic AT &       supervised (multistep) &     fully attention &      large &  800.0 &   6.436619 &        22.0 &      16.0 \\
                  vit\_b,clip,laion2b &        TRADES & self-supervised (multimodal) &     fully attention &      large &  776.0 &   6.614619 &        23.0 &      13.0 \\
                 coatnet\_2,sup,in12k &    Classic AT &                   supervised &              hybrid &      large &  775.0 &   5.877095 &        24.0 &      18.0 \\
convnext\_b,hybrid,laion2b-in12k-in1k &    Classic AT &                       fusion & fully convolutional &      large &  769.0 &   5.440095 &        25.0 &      24.0 \\
                  coatnet\_0,sup,in1k &        TRADES &                   supervised &              hybrid &     medium &  759.0 &   5.377381 &        26.0 &      25.0 \\
               vit\_b,sup,ink22k-in1k &    Classic AT &       supervised (multistep) &     fully attention &      large &  755.0 &   3.658651 &        27.0 &      38.0 \\
            coatnet\_2,sup,in12k-in1k &    Classic AT &       supervised (multistep) &              hybrid &      large &  743.0 &   5.816667 &        28.0 &      20.0 \\
                  vit\_b,clip,laion2b &    Classic AT &                       fusion &     fully attention &      large &  736.0 &   5.730476 &        29.0 &      22.0 \\
                convnext\_b,sup,in22k &    Classic AT &                   supervised & fully convolutional &      large &  729.0 &   5.784381 &        30.0 &      21.0 \\
                 convnext\_b,sup,in1k &        TRADES &                   supervised & fully convolutional &      large &  715.0 &   5.598190 &        31.0 &      23.0 \\
                     swin\_t,sup,in1k &        TRADES &                   supervised &     fully attention &     medium &  713.0 &   5.821698 &        32.0 &      19.0 \\
             convnext\_b,clip,laion2b &    Classic AT & self-supervised (multimodal) & fully convolutional &      large &  692.0 &   3.956905 &        33.0 &      34.0 \\
                   resnet50,sup,in1k &    Classic AT &                   supervised & fully convolutional &     medium &  675.0 &   3.469952 &        34.0 &      40.0 \\
                 coatnet\_2,sup,in12k &        TRADES &                   supervised &              hybrid &      large &  671.0 &   5.048302 &        35.0 &      27.0 \\
                     vit\_b,sup,in22k &    Classic AT &                   supervised &     fully attention &      large &  664.0 &   3.094667 &        36.0 &      47.0 \\
                      vit\_b,sup,in1k &    Classic AT &                   supervised &     fully attention &      large &  643.0 &   2.951619 &        37.0 &      50.0 \\
                 convnext\_t,sup,in1k &        TRADES &                   supervised & fully convolutional &     medium &  639.0 &   4.372286 &        38.0 &      31.0 \\
                  coatnet\_0,sup,in1k &    Classic AT &                   supervised &              hybrid &     medium &  636.0 &   4.020016 &        39.0 &      33.0 \\
           convnext\_b,sup,in22k-in1k &    Classic AT &       supervised (multistep) & fully convolutional &      large &  634.0 &   4.620286 &        40.0 &      29.0 \\
                      vit\_b,sup,in1k &        TRADES &                   supervised &     fully attention &      large &  629.0 &   4.339667 &        41.0 &      32.0 \\
           convnext\_t,sup,in22k-in1k &        TRADES &       supervised (multistep) & fully convolutional &     medium &  619.0 &   3.740952 &        42.0 &      37.0 \\
               mobilenet\_v3,sup,in1k &    Classic AT &                   supervised & fully convolutional &      small &  613.0 &   3.825667 &        43.0 &      36.0 \\
                  vit\_b,clip,laion2b &    Classic AT & self-supervised (multimodal) &     fully attention &      large &  608.0 &   3.844190 &        44.0 &      35.0 \\
               mobilenet\_v3,sup,in1k &        TRADES &                   supervised & fully convolutional &      small &  582.0 &   3.405286 &        45.0 &      41.0 \\
                 edgenetx\_s,sup,in1k &        TRADES &                   supervised &              hybrid &      small &  578.0 &   4.612508 &        46.0 &      30.0 \\
           convnext\_t,sup,in22k-in1k &    Classic AT &       supervised (multistep) & fully convolutional &     medium &  578.0 &   3.261000 &        46.0 &      44.0 \\
                  vit\_b,rob-sup,in1k &    Classic AT &          supervised (robust) &     fully attention &      large &  578.0 &   1.449429 &        46.0 &      68.0 \\
                     vit\_b,dino,in1k &    Classic AT &              self-supervised &     fully attention &     medium &  568.0 &   2.643746 &        49.0 &      57.0 \\
              convnext\_b,clip,laiona &    Classic AT & self-supervised (multimodal) & fully convolutional &      large &  568.0 &   2.857746 &        49.0 &      52.0 \\
                mobilevit\_s,sup,in1k &    Classic AT &                   supervised &              hybrid &      small &  563.0 &   0.316905 &        51.0 &      79.0 \\
                     vit\_s,sup,in22k &    Classic AT &                   supervised &     fully attention &     medium &  550.0 &   3.205714 &        52.0 &      46.0 \\
                mobilevit\_s,sup,in1k &        TRADES &                   supervised &              hybrid &      small &  543.0 &   0.280444 &        53.0 &      80.0 \\
                  vit\_b,rob-sup,in1k &        TRADES &          supervised (robust) &     fully attention &      large &  512.0 &   1.737429 &        54.0 &      65.0 \\
               vit\_s,sup,ink22k-in1k &        TRADES &       supervised (multistep) &     fully attention &     medium &  507.0 &   3.284619 &        55.0 &      43.0 \\
                     swin\_t,sup,in1k &    Classic AT &                   supervised &     fully attention &     medium &  491.0 &   3.246127 &        56.0 &      45.0 \\
                   resnet50,sup,in1k &        TRADES &                   supervised & fully convolutional &     medium &  488.0 &   3.076190 &        57.0 &      48.0 \\
                     deit\_t,sup,in1k &        TRADES &                   supervised &     fully attention &      small &  464.0 &   3.535000 &        58.0 &      39.0 \\
                     vit\_s,sup,in22k &        TRADES &                   supervised &     fully attention &     medium &  449.0 &   2.792762 &        59.0 &      54.0 \\
                 convnext\_b,sup,in1k &    Classic AT &                   supervised & fully convolutional &      large &  447.0 &   2.866635 &        60.0 &      51.0 \\
                regnetx\_004,sup,in1k &    Classic AT &                   supervised & fully convolutional &      small &  432.0 &   2.665524 &        61.0 &      56.0 \\
               vit\_b,sup,ink22k-in1k &        TRADES &       supervised (multistep) &     fully attention &      large &  421.0 &   1.950778 &        62.0 &      63.0 \\
                  eva02\_b,mim,ink22k &        TRADES &              self-supervised &     fully attention &     medium &  409.0 &   2.428794 &        63.0 &      58.0 \\
                convnext\_t,sup,in22k &    Classic AT &                   supervised & fully convolutional &     medium &  400.0 &   1.920302 &        64.0 &      64.0 \\
                     deit\_s,sup,in1k &        TRADES &                   supervised &     fully attention &     medium &  383.0 &   2.965397 &        65.0 &      49.0 \\
                 convnext\_t,sup,in1k &    Classic AT &                   supervised & fully convolutional &     medium &  382.0 &   2.067762 &        66.0 &      61.0 \\
                      vit\_s,sup,in1k &    Classic AT &                   supervised &     fully attention &     medium &  378.0 &   2.048048 &        67.0 &      62.0 \\
                regnetx\_004,sup,in1k &        TRADES &                   supervised & fully convolutional &      small &  375.0 &   2.116175 &        68.0 &      60.0 \\
               vit\_s,sup,ink22k-in1k &    Classic AT &       supervised (multistep) &     fully attention &     medium &  374.0 &   2.748937 &        69.0 &      55.0 \\
                      vit\_s,sup,in1k &        TRADES &                   supervised &     fully attention &     medium &  350.0 &   2.360810 &        70.0 &      59.0 \\
                     deit\_s,sup,in1k &    Classic AT &                   supervised &     fully attention &     medium &  348.0 &   1.705810 &        71.0 &      66.0 \\
                     vit\_b,sup,in22k &        TRADES &                   supervised &     fully attention &      large &  335.0 &   1.039032 &        72.0 &      72.0 \\
                     coat\_t,sup,in1k &        TRADES &                   supervised &              hybrid &      small &  286.0 &   1.462143 &        73.0 &      67.0 \\
                  eva02\_b,mim,ink22k &    Classic AT &              self-supervised &     fully attention &     medium &  279.0 &   1.317603 &        74.0 &      69.0 \\
                     deit\_t,sup,in1k &    Classic AT &                   supervised &     fully attention &      small &  228.0 &   1.072714 &        75.0 &      71.0 \\
                      vit\_b,mae,in1k &        TRADES &              self-supervised &     fully attention &      large &  227.0 &   0.867286 &        76.0 &      74.0 \\
                      vit\_b,mae,in1k &    Classic AT &              self-supervised &     fully attention &      large &  197.0 &   0.688238 &        77.0 &      75.0 \\
                     vit\_b,dino,in1k &        TRADES &              self-supervised &     fully attention &     medium &  187.0 &   0.558254 &        78.0 &      76.0 \\
                 edgenetx\_s,sup,in1k &    Classic AT &                   supervised &              hybrid &      small &  176.0 &   1.225429 &        79.0 &      70.0 \\
                     coat\_t,sup,in1k &    Classic AT &                   supervised &              hybrid &      small &  116.0 &   0.352857 &        80.0 &      78.0 \\
\bottomrule
\end{tabular}
}
\end{table}

\end{document}